\documentclass[10pt,journal]{IEEEtran}
\usepackage{lineno}
\usepackage{times}
\usepackage{epsfig}
\usepackage{graphicx}
\usepackage{amsmath}
\usepackage{amssymb}
\usepackage{booktabs}
\usepackage{multirow}
\usepackage{graphicx}
\usepackage[table,xcdraw]{xcolor}
\usepackage{footmisc}
\newcommand\inv[1]{#1\raisebox{1.15ex}{$\scriptscriptstyle-\!1$}}
\newcommand{\etal}{\textit{et al.}}
\newcommand{\ie}{i.e}
\newenvironment{bluetext}{\par\color{black}}{\par}

\usepackage{amsmath}
\usepackage{booktabs}
\usepackage{multirow}
\usepackage{footmisc}

\ifCLASSOPTIONcompsoc
\usepackage[nocompress]{cite}
\else
\usepackage{cite}
\fi
\ifCLASSINFOpdf
\else
\fi
\begin{document}
		\title{ Backtracking Spatial Pyramid Pooling (SPP)-based Image Classifier for Weakly Supervised Top-down Salient Object Detection}
	
	\author{Hisham Cholakkal, 
		Jubin Johnson, 
		and~Deepu Rajan
		\IEEEcompsocitemizethanks{\IEEEcompsocthanksitem  H. Cholakkal, J. Johnson, and D. Rajan are with the School
			of Computer Science  and  Engineering, Nanyang Technological University, Singapore, 639798. E-mail: \{hisham002, jubin001, asdrajan\}@ntu.edu.sg.\break \vspace{-0.2cm}			
			
			~~This paper has supplementary downloadable material available at https://goo.gl/x6JjSL. The material includes additional experimental results. 
		}
		}
	
	%
	%

	\markboth{IEEE TRANSACTIONS ON IMAGE PROCESSING}%
	{Cholakkal \MakeLowercase{\textit{et al.}}: Weakly supervised top-down salient object detection }
	
	\IEEEtitleabstractindextext{%
		\begin{abstract}
			Top-down saliency models produce a probability map that peaks at target locations specified by a task/goal such as object detection. They are usually trained in a fully supervised setting involving  pixel-level annotations of objects. We propose a weakly supervised top-down saliency framework using only 	binary labels that indicate the presence/absence of an object in an image. First, the probabilistic contribution of each image region  to the confidence of a CNN-based  image classifier is computed through a  backtracking strategy to produce top-down saliency. From a set of saliency maps of an image produced by fast bottom-up saliency approaches, we select the best saliency map suitable for the top-down task. The selected bottom-up saliency map is combined with  the top-down saliency  map. Features having high combined saliency are used to train a linear SVM classifier to estimate  feature saliency. This  is integrated  with  combined saliency and further refined through a multi-scale superpixel-averaging of saliency map. We evaluate the performance of the proposed weakly supervised top-down saliency and achieve comparable performance with fully supervised approaches.  			
			 Experiments are carried out on  seven challenging datasets and quantitative results are compared with 40 closely related approaches across 4 different applications.  
		\end{abstract}
		
		}

		\maketitle
		
		\IEEEdisplaynontitleabstractindextext

		%
		\IEEEpeerreviewmaketitle

\begin{figure}[t]
			\includegraphics[width=1\linewidth, clip=true, trim=0cm 18cm 0cm 0cm]{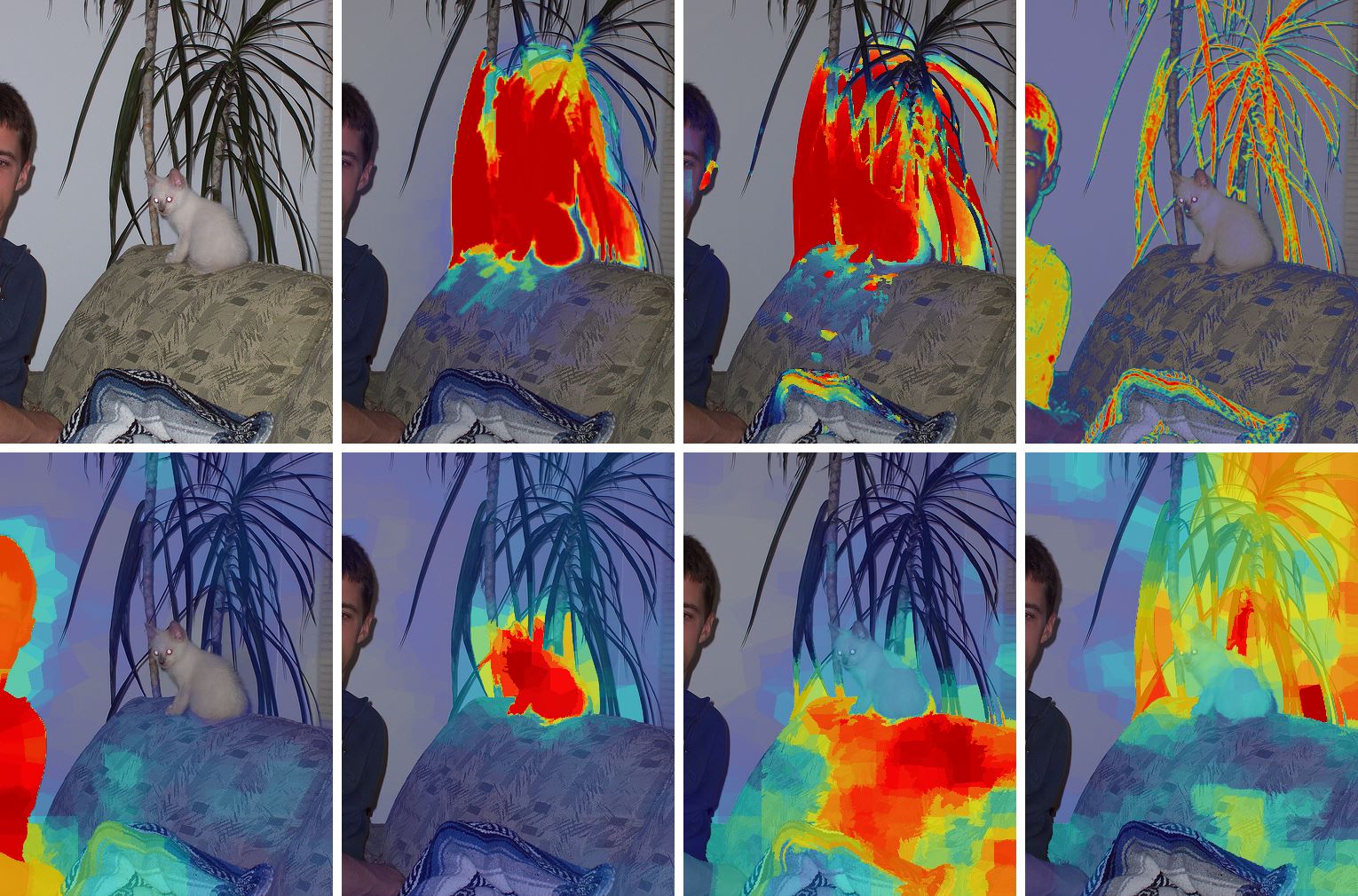}
			\hspace*{0.09\linewidth} (a) \hspace*{0.2\linewidth}(b)\hspace*{0.2\linewidth}(c)  \hspace*{0.19\linewidth} (d)\\  	\vspace{-0.2cm}
			
			\includegraphics[width=1\linewidth, clip=true, trim=0cm 0cm 0cm 18cm]{images_tip/introFigB.jpg}
			\hspace*{0.09\linewidth} (e) \hspace*{0.2\linewidth}(f)\hspace*{0.2\linewidth}(g)  \hspace*{0.19\linewidth} (h)
			 \vspace*{-0.2cm}
			\caption{Comparison of proposed top-down salient object detection with bottom-up methods. (a) Input image, bottom-up saliency maps of (b) MB~\cite{BottomUpICCV2015MDB}, (c)    MST~\cite{bottomup_MST}, and (d) HC~\cite{HC}; proposed top-down saliency maps for (e) person (f) cat (g) sofa and (h) potted plant
				categories.}
			\label{fig:SaliencyMapsVOC2012Intro}
			 \vspace*{-0.6cm}
		\end{figure}

		\ifCLASSOPTIONcompsoc
		\IEEEraisesectionheading{\section{Introduction}\label{sec:introductionCNN}}
		\else
		\section{Introduction}
		\label{sec:introductionCNN}
		\fi
		\IEEEPARstart{T}{he} human visual system has the ability to zero-in rapidly onto salient regions in an image. Recently, there has been much interest among computer vision researchers to model this process known as \textit{visual saliency}, which is attributed to the phenomenon of visual attention. It is beneficial in applications such as object detection/segmentation \cite{R-CNN,Objt_sal_category_indep}, image retargeting \cite{RetargettingSurvey} etc., since identification of salient regions reduces the search space for such high-level tasks. Salient regions in an image are indicated by a probability map called the saliency map. Fig. \ref{fig:SaliencyMapsVOC2012Intro}  shows saliency maps in the form of heat maps, where red indicates higher saliency.
		
		In many instances, the salient region corresponds to a specific object in an image, in which case \textit{salient object detection} becomes a more apt term, wherein pixels belonging to a salient object are assigned high saliency values. Broadly, there are two approaches to salient object detection: bottom-up (BU) \cite{bottomup_MST} and top-down (TD) \cite{JimeiPAMI2016}. The feature contrast at a location plays the central role in BU salient object detection, with no regard to the semantic contents of the scene, although high-level concepts like faces  have been used in conjunction with visual cues like color and shape \cite{liu2011learning}. The assumption that the salient object `pops out'  does not hold when there is little or no contrast between the object and the background. Furthermore, the notion of a salient object is not well-defined in BU models as seen in  Fig.~\ref{fig:SaliencyMapsVOC2012Intro}(b,~c,~d) where BU  methods   \cite{BottomUpICCV2015MDB}, \cite{bottomup_MST} and \cite{HC} show the potted plant in the background as salient to a user searching for the cat.
		
		TD salient object detection is task-oriented and utilizes prior knowledge about the object class. For example, in semantic segmentation \cite{MIL_SemSegWeakCVPR2015}, a pixel is assigned to a particular object class, and a saliency map that aids in this segmentation must invariably be generated by a top-down approach.  
		Fig.~\ref{fig:SaliencyMapsVOC2012Intro}(e, f, g, h)  show the saliency maps produced by the proposed method for  person, cat, sofa and potted plant  categories, respectively. TD saliency is also viewed as a focus-of-attention mechanism by which BU salient points that are unlikely to be part of the object are pruned \cite{topdownDSD}.
		
		Most methods for TD saliency detection learn object classes in a fully supervised manner using pixel-level labeling of objects~\cite{JimeiPAMI2016,topdownBMVC2014,HishamTopdownBMVC2015}. Weakly supervised learning (WSL) alleviates the need for user-intensive annotation by utilizing only class labels for images. Moosmann~\etal~\cite{topdownCategINRIA} propose a weakly supervised TD saliency method for image classification that employs iterative refinement of object hypothesis on a training image. Our method does not require any iterations, yet achieves better results compared to even fully supervised approaches~\cite{ExmSal_CVPR2016,JimeiPAMI2016}. 
		
		  In the preliminary version of this paper presented at CVPR 2016~\cite{HishamCVPR2016}, we  introduced a novel backtracking strategy on multi-scale spatial pyramid max-pooling (SPP)-based image classifier. The objective is to analyze the contribution
		of a feature towards the final classifier score which is then utilized to generate the TD saliency map for an object. 		
		 Following our approach \cite{HishamCVPR2016}, a weakly supervised saliency approach was proposed in \cite{excitationBackPropECCV2016}, which uses  excitation backpropagation to identify the task-relevant neurons in a convolutional neural network (CNN). It produces attention maps that highlight discriminative regions of the object as in \cite{Guidedbackprop2014}, \cite{VisualizeCNN_ICLR2014}, and are only useful for applications such as object localization. The proposed method  highlights the entire object, which enables it to be used in applications such as weakly supervised semantic segmentation and pixel-accurate salient object detection. 
		 We achieve this through the  following novel contributions: 
			
		   (i) A  strategy to backtrack multi-scale spatial pyramid pooled (SPP) CNN features;
			(ii) a feature saliency module that assigns non-zero saliency values at non-discriminative object regions; (iii) a saliency-weighted max-pooling strategy  to select a BU saliency map 
			that is better suited for a given task. 
			 To the best of our knowledge, this is the first work that demonstrates usefulness of top-down salient object detection for a wide range of applications, starting from coarse-level object localization to pixel-accurate semantic segmentation and category-independent salient object detection.

\begin{figure*}[t]
			
			\includegraphics[width=1\linewidth, clip=true, trim= 0.0cm 12.5cm 6.0cm 0cm]{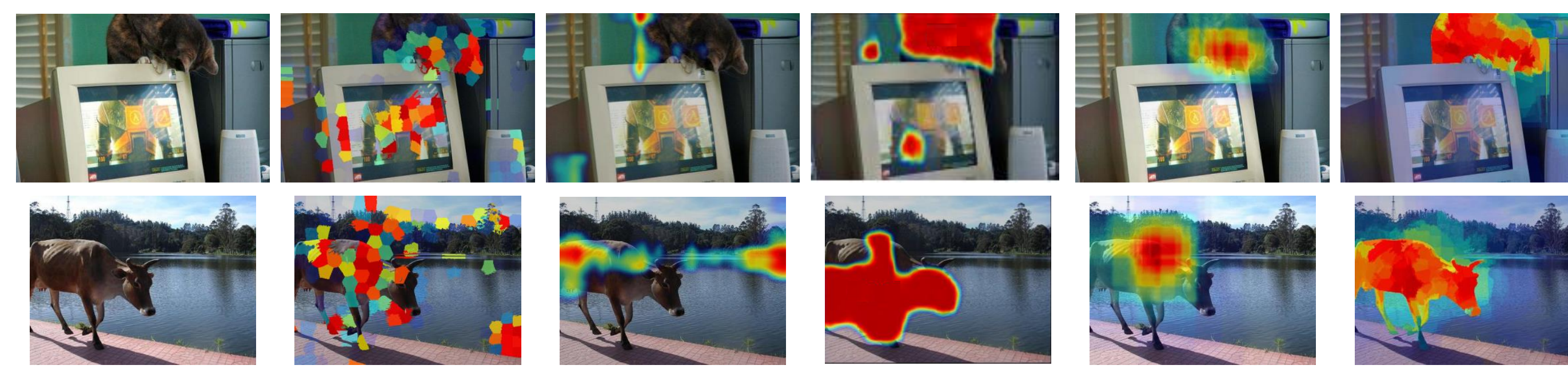}
			\hspace*{0.08\textwidth} (a) \hspace*{0.139\textwidth}(b)\hspace*{0.142\textwidth}(c)  \hspace*{0.138\textwidth} (d) \hspace*{0.133\textwidth} (e)  \hspace*{0.129\textwidth} (f)	\vspace{-0.2cm} 
			
			\caption{Visual comparison of the proposed weakly supervised approach with fully supervised top-down saliency approaches.  (a) Input image,  top-down saliency maps  of (b) Kocak~\etal~\cite{topdownBMVC2014}, (c) LCCSC \cite{HishamTopdownBMVC2015}, (d) Yang and Yang \cite{JimeiPAMI2016}, (e) Exemplar \cite{ExmSal_CVPR2016} and (f)  proposed method for cat (top row) and cow
				(bottom row) categories.} 	
			\label{fig:ComparisonImgFS}
				\vspace{-0.6cm}
		\end{figure*}

			 The major differences in the current version with \cite{HishamCVPR2016} are: 			
					(i) sparse codes of SIFT features in \cite{HishamCVPR2016} are replaced with  CNN features;
					(ii) since  CNN features span larger spatial neighborhood compared to SIFT features, contextual saliency in \cite{HishamCVPR2016} is replaced with  CNN feature  saliency;  
					(iii) \cite{HishamCVPR2016} considers only TD  saliency, while the current version proposes a novel strategy to select a  BU saliency map among several candidates,  which is then combined with TD saliency map; 			
					(iv) multi-scale averaging of saliency values within each superpixel is carried out to improve accuracy along object boundaries; and  	
			 (v) extensive experiments are performed on seven challenging datasets and quantitative results are compared with 40  closely related approaches across 4 different applications.   These modifications lead not only to better performance than \cite{HishamCVPR2016}, but also with fully supervised TD approaches as shown in Fig.~\ref{fig:ComparisonImgFS}.

		We first train a CNN image classifier  using image-level representation of CNN features, that gives a confidence score on the presence of an object in an image. The probabilistic contribution of each  discriminative feature 
		to this confidence score is represented in a TD saliency map, which is combined with a BU saliency map that is selected from several candidate BU maps through a novel selection strategy. 
		Next, the saliency of each feature is separately evaluated using a dedicated feature classifier,  as a means to assign non-zero saliency values to  features from non-discriminative object regions,  based on their  dissimilarity with the background features.
		Saliency inference at a pixel involves combining the image classifier-based saliency map and the feature classifier-based saliency map.
		
		\section{Related Work}
		 We review related work in top-down saliency and relevant applications of CNN under weak supervision. 
	\vspace{-0.4cm}
		\subsection{Top-down saliency frameworks}
			\vspace{-0.05cm}
		
		Kanan~\etal~\cite{topdownSUN} proposed a TD saliency  approach which uses object appearance in conjunction with a location prior. Ineffectiveness of this prior largely affects the  accuracy based on the position of the object within the image.
		Closer to our framework, Yang and Yang~\cite{JimeiPAMI2016} proposed a fully supervised TD saliency model that jointly learns a conditional random field (CRF) and dictionary using sparse codes of SIFT features as latent variables.  
		The inability to discriminate between objects having similar parts (e.g. wheels of car and motorbike) causes a large number of false detections. 
		Kocak~\etal~\cite{topdownBMVC2014} improved upon this by replacing SIFT features with
		the first and second order statistics of color, edge orientation and pixel location within a superpixel, along with objectness~\cite{objectness}. 
		Although this improved the accuracy in distinguishing objects from background, it failed  to discriminate between object categories, causing large number of false detections if the test image contained  objects from other categories, as shown in Fig.~\ref{fig:ComparisonImgFS}(b). 
		Blocking artifacts are also observed in the saliency map at the superpixel boundaries because the superpixels are extracted on a single scale alone. Khan and Tappen~\cite{icip2013discriminative} used label and location-dependent smoothness constraint in a sparse code formulation to produce a smooth saliency map compared to  conventional sparse coding, but with additional computational cost. A joint framework for image classification and TD saliency is proposed in \cite{HishamImageComm2016}. 
		
		Zhu~\etal~\cite{topdown_Contextual} proposed a contextual-pooling based approach where LLC~\cite{LLC_CVPR10} codes of SIFT features are max-pooled in a local neighborhood followed by log-linear model learning. By replacing LLC codes with locality-constrained contextual sparse coding (LCCSC), Cholakkal~\etal~\cite{HishamTopdownBMVC2015} improved  on \cite{topdown_Contextual} with a carefully chosen category-specific dictionary learned from the annotated object area. 
		Discriminative models~\cite{CVPR11_0254,GauravCVPR2012,topdownDSD} often represent a few patches on the object as salient and not the entire object. Hence, such models end up with low recall rates compared to~\cite{topdownBMVC2014,topdownCVPR2012}. 
		In \cite{GauravCVPR2012}, the task of image classification is improved using discriminative spatial saliency to weight visual features.  
		
		In  \cite{ExmSal_CVPR2016}, a fully supervised, CNN-based TD saliency method was proposed that utilized visual association of  query images  with multiple object exemplars. They followed a two-stage deep model  where the first stage  learnt  object-to-object association and the second stage learnt object-to-background discrimination.  Each patch, extracted  using a sliding window, is  
		resized to $ 224\times 224$ and  input  separately to the CNN. 
		There are approximately 500 patches in an image of size $500\times 400$, resulting in 500 forward passes through the network. 
		Training the  model required  more than a week on a GPU.
		Our approach needs only one forward pass to extract CNN features for the entire image, which  reduces the computation  time significantly. It is still able to produce  better saliency maps (Fig.~\ref{fig:ComparisonImgFS}(f)) compared to \cite{ExmSal_CVPR2016} (Fig.~\ref{fig:ComparisonImgFS}(e)).   CNN-based saliency approaches  \cite{DHSNet,MCDL,LEGS,MDF} learn   
		category-independent  salient features \cite{PascalSdataset}   from a large number of fully annotated training images~\cite{HC}.     
		Training or fine-tuning  these saliency models ~\cite{MCDL,MDF} took multiple days, even after initializing their models with  convolutional filter weights pre-trained for  image classification on  ImageNet ~\cite{imageNet}.  
		

		
		The use of weak supervision in TD saliency has largely been left unexamined. Gao~\etal~\cite{topdownDSD} used a weakly supervised setting where bottom-up features are combined with discriminative features that maximize the mutual information to the category label.  In \cite{topdownCategINRIA}, a joint framework using classifier and TD saliency is used for object categorization by sampling representative windows containing the object. Their iterative strategy leads to inaccurate saliency estimation if the initialized windows do not contain the object.
	  In  \cite{VisualizeCNN_ICLR2014,excitationBackPropECCV2016,mahendran16salient}, different variants of CNN backpropagation are used to identify the  image regions  responsible for  activations  corresponding to an object class (salient regions).  
		 	In these discriminative saliency   approaches~\cite{topdownDSD,VisualizeCNN_ICLR2014,excitationBackPropECCV2016}, higher saliency values are assigned only to the features that are discriminative for a category in  an image classification task, limiting their use in applications such as object segmentation, where  all pixels of the object need to be identified accurately. \cite{shimodaECCV16}  improved upon \cite{VisualizeCNN_ICLR2014} by 
		evaluating CNN derivatives with respect to feature maps of the intermediate
		convolutional layer instead of input image. Additionally, they train a fully-connected CRF  for semantic segmentation by using the saliency maps  as unary potentials. We use  the pre-trained  CNN  only as a feature extractor, and the proposed backtracking strategy is applied only from the linear SVM weights to the multi-scale spatial pyramid pooling layer.  Backpropagation of multi-scale spatial pyramid pooling layer is not yet explored for  saliency estimation~\cite{excitationBackPropECCV2016,VisualizeCNN_ICLR2014,shimodaECCV16}. 
		\vspace{-0.3cm}		
		\subsection{CNN-based weakly supervised frameworks}

		Recently, CNN  has been used in a number of weakly supervised object localization approaches \cite{CNN_WS_2014self,LocOquab,CNN_PAMI2016weakly,oquab2014learning}. Multiple-instance learning is applied on CNN features in \cite{CNN_PAMI2016weakly}. In \cite{CNN_WS_2014self}, image regions are masked out to identify regions causing maximal activation.
		The outputs of CNN on multiple overlapping patches are utilized for object localization in  \cite{oquab2014learning}.
		All these approaches need multiple forward passes  on a network to localize objects, which makes them computationally less efficient. 
		Oquab~\etal~\cite{LocOquab} applied global max-pooling to localize a point on objects.  
		Global max-pooling is replaced by average pooling in \cite{MIT_CAM} to help identify the full extent of the object as well.                           
		The underlying assumption is that the loss for average  pooling enables the network to  identify  discriminative object regions. 
		However, the spatial information is lost, whereas it is retained   in our framework via multi-scale spatial pyramid pooling in the image classifier. The image classifier weights are reused  for localization in  \cite{MIT_CAM}.  
		We learn an additional feature classifier to  better estimate saliency at non-discriminative object regions. 
		
		%
		%
		%
		%
		%

		A weakly supervised, end-to-end CNN architecture is  proposed in \cite{ObjDetectCNNBilen}  for  simultaneous object detection and image classification.  Object detection requires classification of  a large number of category-independent object proposals~\cite{SelectiveSearch}, \cite{EdgeProposal}. 
		On a test image, the CNN features  are extracted on the original and flipped image at five scales totaling to 10 feature extraction iterations.   
		In \cite{pronetCVPR2016}, category-specific object proposals  are shown to be effective  for weakly supervised object detection.                               
		Recent semantic segmentation, co-saliency and co-segmentation frameworks   \cite{SemanticSeg_pathak,MIL_SemSegWeakCVPR2015,Coseg_Quan_2016_CVPR,CoSalIJCV016}  also train their CNN models in a weakly supervised setting.
		
		Internal representations learned by CNN are visualized in \cite{VisualizeCNN_ECCV2014,VisCNN_Invert_CVPR15,VisCNN_2015inverting,VisCNN_ObjectDet,Guidedbackprop2014,mahendran16salient}  for better understanding of its properties. \cite{VisualizeCNN_ECCV2014} and \cite{VisCNN_ObjectDet} analyze the convolution layers using techniques such as  deconvolutional networks. In \cite{VisCNN_Invert_CVPR15} and \cite{VisCNN_2015inverting}, CNN features are inverted at different layers of the network  including the fully connected layers, to analyze the visual encoding of CNN. 
		
		%
	    
				\vspace{-0.1cm}	
		\section{Proposed method}
				\begin{figure*}[t]
					\centering
					\includegraphics[width=0.9\linewidth, clip=true, trim=0.0cm 8.5cm 0.0cm 0.2cm]{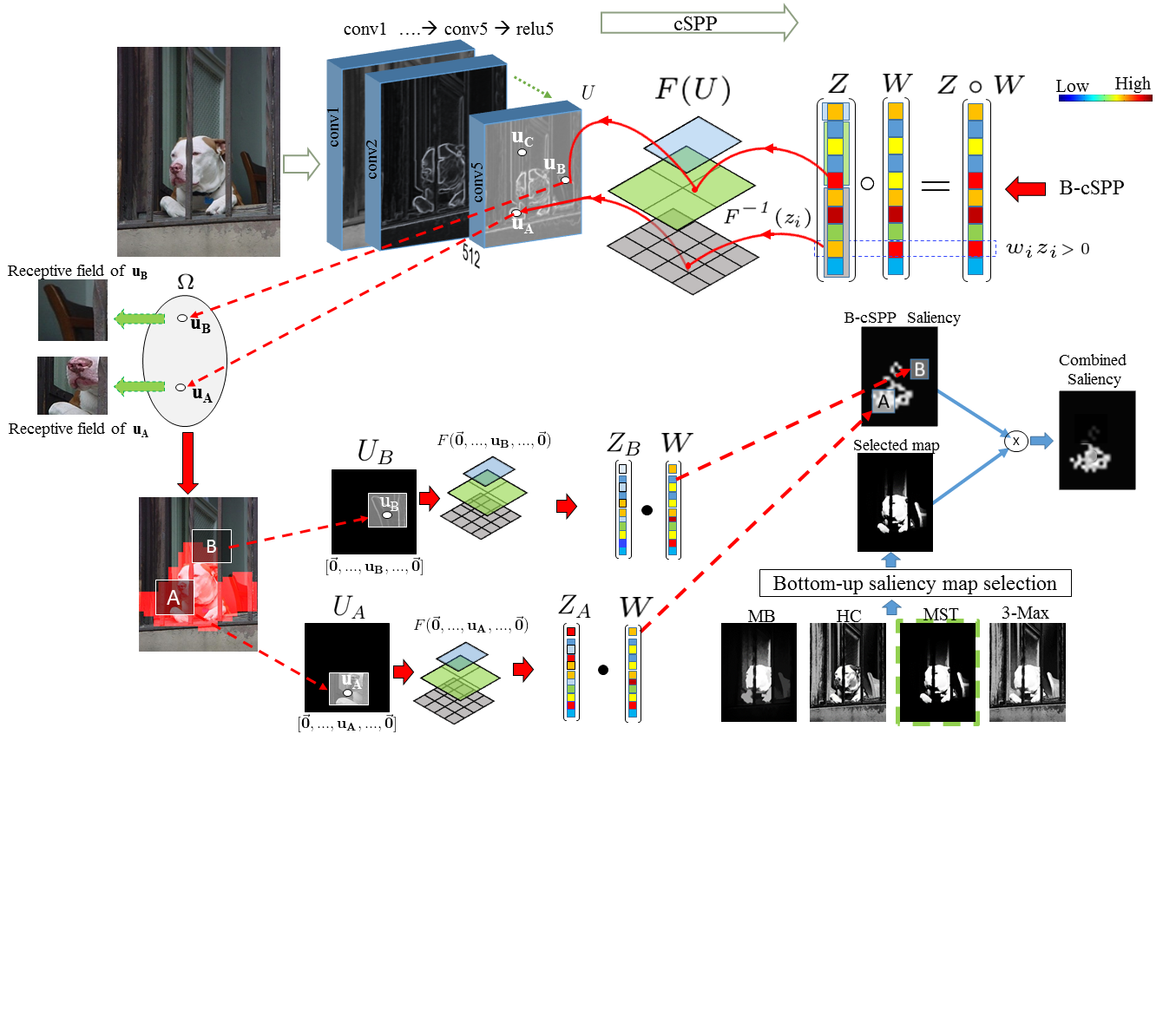}
					\caption{Illustration of combined saliency estimation for \textit{dog} category.  Red arrows indicate the proposed backtracking strategy for top-down saliency (B-cSPP). From a set of BU saliency maps (MB~\cite{BottomUpICCV2015MDB}, HC~\cite{HC},  MST~\cite{bottomup_MST}  and 3-Max),
						 the best one is selected and is integrated with B-cSPP saliency to produce combined saliency. 3-Max is the saliency map obtained by taking the maximum saliency at each pixel across the 3 BU saliency maps.}
					\label{fig:R_ConvSPM}
						\vspace{-0.25cm}
				\end{figure*}
	  	 In this section, we first describe the CNN image classifier and how the backtracking mechanism generates the TD saliency map. Next, a novel strategy to select a BU map is described followed by its integration with the TD saliency map  to obtain combined saliency.  A dedicated feature saliency model is learned  on  features extracted using this combined saliency to evaluate saliency at each feature independently.  Finally, inference  involves combining the feature saliency with the combined saliency to obtain the TD  map. 

				\vspace{-0.1cm}	
		\subsection{CNN  Image Classifier} 
		\label{sec:ConvSPM}
		%
		%
		The CNN features of an image are extracted using VGG-16~\cite{VGG_VD16}   that has been pre-trained on the ImageNet ILSVRC 2012 data~\cite{imageNet} with only image-level supervision (no bounding box annotation).  All the layers of VGG-16 upto relu5\_3  (third rectified linear unit  in the fifth layer)  are used for feature extraction and the convolution weights pre-trained  for image classification are used without any fine-tuning. However, unlike  \cite{VGG_VD16,VGGnet}, we do not crop/resize the input image for feature extraction, but use the original image at its full resolution.
		A fixed-length, image-level representation of  CNN features is obtained through a multi-scale spatial pyramid max-pooling as in Spatial Pyramid Pooling (SPP-net) \cite{SPPnet} image classifier. We use  a binary linear SVM after the spatial pyramid pooling layer, instead of  fully connected layers in  \cite{SPPnet}.  In order to reduce computations, we extract  CNN features at a single image scale instead of multiple scales.
		
%
%

		The spatial dimensions of an input image are down sampled by  a factor of 16  at the relu5\_3 feature map due to spatial max-pooling in the first four layers. 
				 There are  $d$ (=512) convolution filters in conv5\_3 (third convolution in the fifth layer) which results in  $d$ activation maps at the output of succeeding relu5\_3.  
				We consider the activation at each spatial location of relu5\_3 activation maps as a  $d$ dimensional feature vector, which we refer  to as relu5 feature. The relu5 feature represents the overall response of multiple pixels from its receptive field in the original image.  

	 	Let $U=[\mathbf{u_1}, \mathbf{u_2}...\mathbf{u_m},...\mathbf{u_M}]$ denote  $M$ relu5 features each of dimension $d$. 
		The spatial distribution of the features in the image is encoded in the spatial pyramid  max-pooled image vector $Z$ through a multi-scale max-pooling operation $F(\mathbf{u_1}, \mathbf{u_2},...\mathbf{u_M})$ of the relu5 features  on a 3-level spatial pyramid~\cite{SPM_CVPR_06} as shown in Fig.~\ref{fig:R_ConvSPM}. The $i^{th}$ element $z_{i}$ of $Z$ is a max-pooled value derived using maximum operation on $j^{th}$ element of all relu5 features in a spatial pyramid region $\mathcal{R}$ defined by $i$, and $ j=\{1+ (i-1)~{mod}~d\}$. \ie,  \begin{bluetext}  		
	     	\begin{align*}			
			&\begin{aligned}[t]
			\hspace{-3.2cm} 
			z_i=F(\mathbf{u_{1j}}, \mathbf{u_{2j}}...\mathbf{u_{mj}},...\mathbf{u_{Mj}}) 		
			\end{aligned}
				\end{align*} 
					\begin{align}
			&\begin{aligned}[t]
			&~~~ = max\{u_{1j}, u_{2j}, ....u_{qj}\},   \forall \;  \mathbf{u_{1}},\mathbf{u_{2}}...\mathbf{u_{q}} \in  \mathcal{R}.
			\end{aligned}		  
		 	\end{align}  
		Let the label $Y_k\in\left \{1,-1 \right \}$ indicate the presence or absence of an object   $O$  in the $k^{th}$ image. If $Y_k=1$, it is a positive image, else it is a negative image.  Image-label pairs ($Z_k, Y_k$) of $T$ training images are used to train a  binary  linear SVM classifier~\cite{vlfeat,SVM_SDCA} that minimizes the following loss function, 
		   \begin{equation}
		   \label{eq:svm_eq}
		      \underset{W}{\arg\min} \frac{\lambda}{2} \|W\|^2 +  \frac{1}{T}\displaystyle\sum_{k=1}^T\ max(0,1-Y_k (W^\top Z_k+ bias)) ,
		    \end{equation}
		  where  $W=[w_{1}, w_{2}....w_{N}]^\top$ and $bias$ are the SVM weight vector and  bias respectively and  are learnt  for each object category.     
		$N$ is the length of the max-pooled image vector $Z_k$  and $\lambda$ is a regularization constant empirically set to 0.01 through cross validations.\end{bluetext}
		
		Given a validation/test image with max-pooled vector $Z$, the classifier score $ {W}^\top {Z}+b$ indicates the confidence of the presence of object $O$ in it. 
		Our CNN image classifier is referred as cSPP in the following discussions.
		
		%
		
			\vspace{-0.25cm}	
		\subsection{Backtracking image classifier for saliency estimation}
			\vspace{-0.05cm}
		\label{sec:RConvSPM}
		In the cSPP image classifier, both the linear-SVM and multi-scale max-pooling operations can be traced back to the relu5 feature locations. This enables us to analyze the contribution of each feature towards the final classifier score which is then utilized to generate the  TD saliency map for an object. 
		
		First, we estimate the ability of a  relu5  feature to represent a spatial pyramid region $\mathcal{R}$ by counting the number of times the elements of that feature made it to the spatial pyramid max-pooled vector.
		We call this the representativeness, $r_m$, of a feature $m$. 
		Representative features may either contribute positively or negatively to the classifier score with higher contribution indicating more relevance of the feature to an object $O$. The relevance of the feature  to the object is denoted $c_m$. 
		
		For a positively contributing relu5 feature, it is possible that there are some negatively contributing elements among the total $d$ elements.  
		 For example, let $\mathbf{u_m}=[u_{m1},0,...\;u_{mj}\;...0,u_{md}]^\top$ be the $m^{th}$  feature with its $j^{th}$ element $u_{mj}$ being a local maximum in its spatial pyramid region. Although $u_{mj}$ contributes positively to the classifier confidence ${W}^\top Z+b$, the other non-zero elements $u_{m1}$ or $u_{md}$ may contribute negatively, indicating absence of the object. So, the relevance of a feature to the object requires its contribution to be computed in the absence of other features; this relevance is denoted $p_m$.  We define the top-down saliency $S_\mathcal{T}(\mathbf{u_m})$  of the  feature $\mathbf{u_m}$ to the object class $O$ as the joint probability of three parameters - $r_m, c_m$ and $p_m$ as 
		\begin{equation}
			\vspace{-0.05cm}
	P(p_m,r_m,c_m)= P(p_m|r_m,c_m)P(c_m|r_m)P(r_m).
		\label{eq:psaliency}
			\vspace{-0.05cm}
		\end{equation}  
		
		The representative elements of the  feature  $\mathbf{u_m}$ are identified as
		\begin{equation}
			\vspace{-0.05cm}
		\Psi_m=\{i\delta(\inv{F}(z_i),u_{mj})\}, \quad \forall i \in \{1, 2,..N\},
			\vspace{-0.02cm}
		\end{equation}  
		  \begin{bluetext} where  $\delta$  is the Kronecker delta function and  the location of $z_i$ in $Z$  identifies the region $\mathcal{R}$ in the spatial pyramid and its position $j$ in the relu5 feature  $\mathbf{u_m}$. $\inv{F}$ is the inverse operation of multi-scale spatial pyramid max-pooling illustrated in Fig.~\ref{fig:R_ConvSPM} and written as 
		  	  	\begin{equation}
		  	  	\inv{F}(z_i)=
		  	  	\begin{cases}
		  	  	u_{nj},~~~~ 
		  	  	\;\;  \text{if}~u_{nj}=z_i,  \forall \;  \mathbf{u_{n}} \in  \mathcal{R}\\
		  	  	0 \quad \quad\quad \;\;\;\text{otherwise.}
		  	  	\end{cases}
		  	  	\label{eq:Eqimplementation}	
		  	  	\end{equation}  		  	  
		  \end{bluetext}
		   The probability of representativeness of the $m^{th}$ feature to the image is then defined as 
		\begin{equation}
			\vspace{-0.05cm}
		P(r_m)= \dfrac{\mathbf{card}(\Psi_m)}
		{		
		\tiny{\sum_{\forall i \in \{1,..M\}}} \mathbf{card}(\Psi_i)},		
		\label{eq:prm}
			\vspace{-0.0cm}
		\end{equation}
		where $\mathbf{card}(.)$  is the cardinality.
		
		The linear SVM classifier confidence is a score indicating the presence of the object in the image, which  increases from a definite absence ($score\leq-1$) to definite presence ($ score\geq 1$). The confidence of $Y=1$ is 
		\vspace{-0.05cm}
		\begin{align*}			
		&\begin{aligned}[t]
		\hspace{-1cm} 
		\Theta~(Y=1\,|\,F(\mathbf{u_1},\mathbf{u_2},..., \mathbf{u_M}))
		\end{aligned}\\
		&\begin{aligned}[t]
		&\hspace{-0.05cm} =  W^\top F(\mathbf{u_1}, \mathbf{u_2},..., \mathbf{u_M}) +b,
		\end{aligned}\\
		&= \begin{aligned}[t]
		W^\top Z +b  \;\; =  \hspace{-0.3cm}\sum_{\forall i \in \{1,..N\}}\hspace{-0.5cm} w_iz_i +b,
		\end{aligned}    \\  
		&= \begin{aligned}[t]
		\hspace{-0.2cm} \sum_{\forall i \in\Psi_m} \hspace{-0.2cm} w_iz_i \hspace{0.2cm}  +\hspace{0.3cm}   \hspace{-0.7cm} \sum_{\forall i\in \{1,..N\} \setminus\Psi_m}\hspace{-0.7cm} w_iz_i+b,
		\end{aligned}   \\   
		&= \begin{aligned}[t]
		\theta(c_m|r_m) \hspace{0.2cm}  +\hspace{0.3cm} \hspace{-0.7cm} \sum_{\forall i\in \{1,..N\} \setminus\Psi_m}\hspace{-0.7cm} w_iz_i+b,
		\end{aligned}  
		\end{align*}
		where  $\theta(c_m|r_m)$ is the contribution of $m^{th}$  feature $\mathbf{u_m}$ to the image classifier confidence and $N$ is the length of $Z$.
		
		Given that the feature  is representative of the image, the probability of it belonging to the object  $O$  is 
		\begin{equation}
		P(c_m|r_m)= 
		\begin{cases}
		\beta~(\theta(c_m|r_m) ), \quad \text{if}  \quad\theta(c_m|r_m) \geq 0,\\
		0, \quad \;\quad \quad \quad \text{otherwise}.
		\end{cases}
		\label{eq:pcmrm}
			\vspace{-0.05cm}
		\end{equation}
		  where $\beta$ is the sigmoid function that maps the confidence score to [0,\,1]  and ensures that the probabilities  of $\mathbf{u_m}$ belonging to object $O$ and not belonging to object $O$ sums to 1.
		
		Using the above probabilities, we select a set $\Omega$ of all features that contribute positively to the classifier confidence as 
		\vspace{-0.25cm} 
		\begin{equation}			
		\begin{split}
		\Omega=\{ P(c_m|r_m) P(r_m)>0 \}, \;\forall m=1,2,...,~M.
		\end{split}
			\vspace{-0.05cm}
		\end{equation}
		
		The net contribution of a feature $\mathbf{u_m} \in \Omega$ in the absence of other features is 
		\begin{equation}
		P(p_m|r_m,c_m)= \beta~(W^\top F(\vec{\mathbf{0}} .., \mathbf{u_m},..., \vec{\mathbf{0}} ) +b),
		\label{eq:ppmrmcm}
		\end{equation}	
		where $F(\vec{\mathbf{0}} .., \mathbf{u_m},..., \vec{\mathbf{0}} )$ is the spatial pyramid  max-pooling operation performed by replacing all features except $\mathbf{u_m}$  with a zero vector $\vec{\mathbf{0}} $ of size $d$ to form  max-pooled vector $Z_m$. 
		
		%
		\textbf{Implementation.}
		Fig.~\ref{fig:R_ConvSPM} illustrates three relu5 features $\mathbf{u_A}$, $\mathbf{u_B}$ and $\mathbf{u_C}$. 
		The confidence of the presence of  object $O$ in an image is indicated by the classifier score $ {W}^\top Z+b$  as mentioned in the previous section  and the element $z_i$ of $Z$ has a corresponding weight $w_i $. The  elements from the Hadamard product $W \circ Z$ with  $w_i z_i>0$  mark
		the features $\mathbf{u_A}$ and $\mathbf{u_B}$ that contribute positively to the classifier confidence through a $\inv{F}(.)$ operation,  i.e the set $\Omega$. The contribution of  feature $\mathbf{u_A}$ in the absence of other features is evaluated using max-pooling operations $F(\vec{\mathbf{0}}.., \mathbf{u_A},..,\vec{\mathbf{0}} )$   in which   all  features except $\mathbf{u_A}$ are replaced with $\vec{\mathbf{0}}$ forming max-pooled vector $Z_A$.  
		The  saliency of a feature $m$ is given by
		\begin{equation}
		S_\mathcal{T}(\mathbf{u_m})=
		\begin{cases}
		\beta~ (W^\top F(\vec{\mathbf{0}}.., \mathbf{u_m},..,\vec{\mathbf{0}} )+b ) \;\;  \text{if}~ m \in \Omega,\\
		0 \quad \quad\quad \quad\quad\quad\quad\quad\quad\quad\quad\;\;\;\text{otherwise.}
		\end{cases}
		\label{eq:Eqimplementation}	
		\end{equation} 
				
		Since this TD saliency of a feature is arrived at by backtracking the cSPP classifier, we call it \textit{B-cSPP saliency} and the corresponding saliency map as \textit{B-cSPP saliency map}. The feature $\mathbf{u_A}$  from the object (dog) region  is assigned  high B-cSPP saliency while  $\mathbf{u_B}$  from background is assigned  zero B-cSPP saliency. 
		
		\subsection{Selection of bottom-up saliency map} 
				\vspace{-0.1cm}
		\label{BottomUpSelection}
		 \begin{bluetext} 	
		 	Contextual background regions help in improving image classification tasks \cite{excitationBackPropECCV2016,VisualizeCNN_ICLR2014,shimodaECCV16}.   
		 	For example, grass or sky in the background can help in classifying cow or bird category,  respectively. 		   Salient object detection should exclude such background regions. Bottom-up saliency assigns low saliency to these  uniform  regions that have low contrast with the surroundings. 
		    Hence, we multiply the B-cSPP saliency
		 	with a bottom-up saliency selected through a novel selection
		 	strategy, to  form a combined saliency. 
		 	In this section, we explain our  novel strategy to select a saliency map which is best suited for the task at hand, from a set of BU maps based on a saliency weighted max-pooling.  

		 \end{bluetext}
		 
		
		State-of-the-art BU saliency approaches~\cite{BottomUpICCV2015MDB,bottomup_MST} can produce a category-independent saliency map for an image within 40  milliseconds. They assume  image boundaries as the background while approaches  such as~\cite{HC} focus on feature contrast to estimate  saliency. These approaches do not require any training and give reasonably good results. Since BU saliency maps are task-independent from a user's perspective, the definition of `good saliency map' varies based on the application. For example, consider  Fig.~\ref{fig:SaliencyMapsVOC2012Intro}, where   four different objects are present. If a user searches for  a `person' in the image,  BU approaches~\cite{BottomUpICCV2015MDB,bottomup_MST}  that  assume  image boundary as the background fail to produce  a `good saliency map'. In such scenarios, an approach~\cite{HC}  that does not use such assumptions can produce better results. Thus, our objective is to develop a strategy to select a BU saliency method for a particular image that is best suited for the task at hand.

		Our cSPP  image classifier $(W,b)$ which was trained to estimate the presence of object $O$ in an image is employed to select a  BU saliency map suitable for the task of identifying  image regions that belong to object $O$. To achieve a one-to-one correspondence between  pixels in the BU saliency map and  the relu5 features, we downsample  the saliency maps  to the spatial resolution of feature map at relu5\_3, \ie, by a factor of 16. 
		%
		%
		From  $n_\rho$ BU saliency maps, we need to select  one for which features that belong to an object  are assigned high saliency and those that do not belong to an object are assigned low saliency.
		%
		For a max-pooled vector $Z$ of an image, the SVM predicts a confidence score  $ {W}^\top {Z}+b$ which is proportional to the confidence of object presence  in that image. \ie,
		\vspace{-0.15cm}
		\begin{align*}
		&\begin{aligned}[t]
		\hspace{-1.5cm}
		\Theta~(Y=1\,|\,Z)
		\end{aligned}\\
		&= \begin{aligned}[t]
		W^\top Z +b =  \hspace{-0.4cm}\sum_{\forall i \in \{1,..N\}}\hspace{-0.5cm} w_iz_i +b,
		\end{aligned}    \\  
		&= \begin{aligned}[t]
		\hspace{-0.2cm} \sum_{\forall i \in \;\mathcal{I^+}} \hspace{-0.2cm} w_{i}z_{i}\hspace{0.0cm}  +\hspace{0.3cm}   \hspace{-0.1cm} \sum_{\forall i \in \mathcal{I^-}}\hspace{-0.0cm} w_iz_i +b.
		\end{aligned}  
		%
		\end{align*}
		\vspace{-0.05cm}
		\begin{equation}
		\label{eq:svmConf}
		\Theta~(Y=1\,|\,Z)=\hspace{-0.2cm} \sum_{\forall i \in \mathcal{I^+}} \hspace{-0.2cm} w_{i}z_{i} \hspace{-0.1cm}  -\hspace{-0.0cm}   \hspace{-0.15cm} \sum_{\forall i \in \mathcal{I^-}}\hspace{-0.0cm} |w_i|z_i +b,
		\end{equation} 
		\vspace{-0.5cm}
		where 
		\begin{equation*}
		\mathcal{I^+}=\{ i\; |\; w_i> 0\}, \quad \forall i \in \{1, 2,..N\},
		\end{equation*} 
		\begin{equation*}
		\mathcal{I^-}=\{ i\; |\; w_i< 0\}, \quad \forall i \in \{1, 2,..N\}.
		\end{equation*}

		Ideally, features  belonging to object $O$ 
		contribute positively to the classifier confidence and hence they correspond to  elements in $Z$ whose indices belong to $\mathcal{I^+}$,  while the background features result in  $\mathcal{I^-}$ indices. It is to be noted that $z_i$ is non-negative since it is derived from relu5 through max-pooling operation. 
		
		First, the $m^{th}$ feature $\mathbf{u_m}$ is weighted with  $\rho_m^t$, the BU saliency value for that feature estimated by $t^{th}$ approach. i.e, $\mathbf{\hat u_m} = \mathbf{u_m} \times \rho_m^t $.  
		The saliency-weighted relu5 features  $\hat U=[ \mathbf{\hat u_1}, \mathbf{\hat u_2}...,...\mathbf{\hat u_M}]$  are used to estimate the saliency-weighted max-pooled vector $\hat Z$ and similar to  Eq.~(\ref{eq:svmConf}), the modified confidence score $\hat{B}(t)=\Theta~(Y=1\,|\,\hat Z)$ due to the  $t^{th}$ BU  map is computed as,  
		\begin{equation}		
		\vspace{-0.15cm}
		\hat{B}(t)= \hspace{-0.2cm} \sum_{\forall i \in \mathcal{I^+}} \hspace{-0.2cm} w_{i} \hat z_{i} \hspace{-0.1cm}  -\hspace{-0.0cm}   \hspace{-0.15cm} \sum_{\forall i \in \mathcal{I^-}}\hspace{-0.0cm} |w_i| \hat z_i +b.
		\label{eq:positSalWa}
		\end{equation} 
		If higher values in the saliency map produced by algorithm $t$ falls exactly on the object regions,  the  second summation will be largely reduced, due to weighting background indices with low saliency values and hence   $\hat{B}(t)$ will be high. If some of the background also garners high saliency, then $\hat{B}$ will be relatively low. 
		 In order to reinforce the above assertion, we invert the saliency map  (by subtracting saliency values from the maximum saliency value in the image), and  recompute  the saliency-weighted relu5 features ($\tilde{U}$), saliency-weighted max-pooled vector $\tilde{Z}$,  and $\tilde{B}(t)$.
						
				    Let   $\rho^t_{max}$ be the  maximum  value in the saliency map of  an image by $t^{th}$ approach, then  the inverted saliency value for $m^{th}$ location is   
					\begin{equation}
					\label{eq:svmConf}
					\tilde{\rho}_m^t=\rho^t_{max}-\rho_m^t,
					\end{equation} 				
					Then the $m^{th}$ feature $\mathbf{u_m}$ is weighted with  $\tilde{\rho}_m^t$, resulting in $\mathbf{\tilde u_m}$, i.e., 
					\begin{equation}
					\mathbf{\tilde u_m} = \mathbf{u_m} \times \tilde{\rho}_m^t.
					\end{equation}\textcolor{black}{Repeating the above procedure for all relu5 features we obtain $\tilde{U}$, the  relu5 features  weighted with inverted saliency map.
					Multi-scale spatial pyramid pooling ($F(.)$ ) of $\tilde{U}$ will result in saliency-weighted max-pooled vector $\tilde{Z}$. i.e,
					\begin{equation}
					\tilde{z}_i=F(\mathbf{\tilde{u}_{1j}}, \mathbf{\tilde{u}_{2j}}...\mathbf{\tilde{u}_{mj}},...\mathbf{\tilde{u}_{Mj}}) 		
					\end{equation} }  	
				\textcolor{black}{Note that $F(.)$ is a non-linear operation, hence $\hat{z}_i \neq \rho^t_{max}- \tilde{z}_i $. Finally, $\tilde{B}(t)$ is} 							
		\begin{equation}
		\tilde{B}(t)= \hspace{-0.2cm} \sum_{\forall i \in \mathcal{I^+}} \hspace{-0.2cm} w_{i}  \tilde {z}_{i} \hspace{-0.1cm}  -\hspace{-0.0cm}   \hspace{-0.15cm} \sum_{\forall i \in \mathcal{I^-}}\hspace{-0.0cm} |w_i| \tilde {z}_i +b.
		\label{eq:positSalWeight2}
		\end{equation} 
		If all object regions are assigned with higher saliency values  in Eq.~(\ref{eq:positSalWeight2}), higher weights are assigned to the background regions and lower weights to the salient regions, leading to a lower score of  $\tilde{B}(t)$.   
		Combining the above two observations, an ideal saliency map should maximize  
		\begin{equation}
		\hat{B}(t)-\tilde{B}(t)= \hspace{-0.2cm} \sum_{\forall i \in \mathcal{I^+}} \hspace{-0.2cm} w_{i} (\hat z_{i}-\tilde{z}_{i})  \hspace{-0.0cm}  -\hspace{-0.0cm}   \hspace{-0.15cm} \sum_{\forall i \in \mathcal{I^-}}\hspace{-0.2cm} |w_i|\hspace{0.1cm}(\hat z_i-\tilde{z}_{i}).
		\label{eq:positSalWeight3}
		\end{equation} 
				
		In order to prevent the selection of a map that assigns high saliency to the entire image, we impose a penalty of $1-\mu_{t}$  on saliency map  $t$ with a mean  saliency $\mu_{t}$. Combining the above observations, the final objective function to select a BU saliency map  is 		
		\begin{equation}
		\mathcal{B}(t)= \hspace{-0.0cm} \{\sum_{\forall i \in \mathcal{I^+}} \hspace{-0.2cm} w_{i} (\hat z_{i}-\tilde{z}_{i})  \hspace{-0.0cm}  -\hspace{-0.0cm}   \hspace{-0.15cm} \sum_{\forall i \in \mathcal{I^-}}\hspace{-0.2cm} |w_i|\hspace{0.1cm}(\hat z_i-\tilde{z}_{i}) \}\times (1-\mu_{t}).
		\label{eq:positSalWeight4}
		\end{equation} 
		If the saliency map of $t^{th}$ algorithm is not aligned  with the object, then the false positives will increase $\hat z_i $  and decrease $\tilde{z} $ in $\mathcal{I^-}$, thus increasing the second term of Eq.~(\ref{eq:positSalWeight4}). False negatives will reduce $\hat z_i$  and increase  $\tilde{z}$   reducing 
		the first term. 
		Hence an inaccurate BU saliency map will result in low $\mathcal{B}(t)$. The saliency map that maximizes Eq.~(\ref{eq:positSalWeight4})  is selected.  In right side of   eqs.~(\ref{eq:positSalWeight2}), (\ref{eq:positSalWeight3}) and (\ref{eq:positSalWeight4}), the saliency map of  $t^{th}$ algorithm is implicit in $\hat z_i$ and $\tilde{z_i} $ through $\mathbf{\hat u_m}$ and $\mathbf{\tilde u_m}$.


		In addition to choosing individual BU saliency maps, we also analyze whether a combination of these maps has an effect on improving TD saliency. To this end, we combine saliency maps by picking the maximum saliency for each pixel and use Eq.~(\ref{eq:positSalWeight4}) to select the best map from a set of saliency maps that includes the maximum map. In this section, we have assumed that the SVM weights learnt for an object is accurate and that the object appears only at locations where $w_i$ are positive. Although this may not be always true, we retain this assumption since object locations are not available in a weakly supervised setting.
		
		Off-the-shelf BU  methods can be used to generate the BU saliency maps.
        We used  MB~\cite{BottomUpICCV2015MDB}, MST~\cite{bottomup_MST}  and HC~\cite{HC}  in our framework to  improve the runtime performance.
		 The B-cSPP saliency map and the selected bottom up saliency map are combined through a simple multiplication as shown in Fig.~\ref{fig:R_ConvSPM}.  We denote this combined saliency map as $\mathcal{H}$.
		Following \cite{JimeiPAMI2016,topdownDSD,topdownBMVC2014}, we also characterize our category-specific saliency inference framework as TD saliency even though there is a bottom-up component.
		 \vspace{-0.35cm}
		\subsection{Feature saliency training}
		 \vspace{-0.1cm}
		\label{sec:contextual_saliencyCNN}
		
		Image classifiers trained on image-level representation of features  have shown to be effective in discriminative TD saliency estimation \cite{HishamCVPR2016,GauravCVPR2012,topdownDSD}. The combined saliency map $\mathcal H$ takes non-zero values only at discriminative image regions whose features  make positive contribution to the image classifier confidence. 
		The assumption is that the object appears only at
		grids in the spatial pyramid where $w_i$ are positive, which may not be true across all images. Our objective is not limited to identifying the discriminative image regions, but to assign higher saliency values to all pixels belonging  to the salient object.  
		In order to independently estimate the saliency value of each  relu5 feature, we also learn a top-down \textit{feature saliency} model that uses a linear SVM learnt on positive and negative relu5 features from the training images. Since feature-level annotation is not available, we  use object features extracted using the combined saliency map $\mathcal H$  to train the model.
		
		
		From positive training images of object $O$, relu5 features with $\mathcal{H}$ saliency greater than $0.5$ are selected as positive features with label $l=+1$. In order to prevent training features from non-discriminative object regions of positive images with negative label, only those features at which both B-cSSP and BU saliency are selected as negative features with label $l=-1$.  
		Additionally,  random features are selected from negative images with  label $l=-1$. 
		A linear SVM model with weight $\mathbf{v}$ and bias $b_v$ is learned. 
		Since the relu5 features are already computed for B-cSPP,  learning of  linear SVM is the only additional computation required to train this top-down model. The saliency map obtained from feature saliency is denoted~$\mathcal L$.
		  \vspace{-0.44cm}
		\subsection{Saliency inference}
		 \vspace{-0.08cm}
		\label{sec:saliency_inferenceCNN}
		
		For inference on a test image, the combined saliency $\mathcal{H}$ and feature saliency are first integrated followed by multi-scale superpixel averaging and finally associated with the confidence of the image classifier to obtain the saliency at a pixel. While the combined  saliency is obtained as described in Section~\ref{BottomUpSelection}, the feature saliency for a feature $\mathbf{u_m}$ is the probability of the feature belonging to an object 
		computed by applying a sigmod function $\beta$ to the linear SVM score,  
		\begin{equation}
		P(l = 1\mid \mathbf{u_m} ,\mathbf{v}) = \beta~(\mathbf{v}^T \mathbf{u_m} + b_v).
		\label{eq:regression}
		\end{equation}
		The feature saliency and combined saliency values are integrated using a mean operation to form the saliency map,  $S_p= \frac{\mathcal H+ \mathcal L}{2}$.
	
		\subsubsection{Multi-scale superpixel-averaging of saliency map}
		\label{sec:saliencyPropCNN}
		
		The low resolution saliency map $S_p$ is upsampled to the original image size  using bicubic interpolation. 
		As a consequence, saliency values may not be uniform within a superpixel. Also, the saliency map will not be edge-aware with 
		object regions spreading to the background.
		Hence, a  multi-scale superpixel-averaging strategy is employed.
		The mean saliency at a superpixel (obtained by SLIC segmentation ~\cite{SLIC}) is assigned to every pixel in it. This process is repeated at multiple scales by varying the SLIC parameters.
		The resulting maps are averaged to produce a smooth, pixel-level saliency map $S_{pix}$ that uniformly highlights the salient object  and also produces a sharp transition at object boundaries.
		\subsubsection{Integrating with image classifier confidence}
		For a given image, the TD saliency  map $S_{pix}$ indicates the probable pixels that belong to object $O$. 
		Since the presence of a specific object  in a test image is  not known apriori for applications such as semantic segmentation and object detection,  the  saliency map needs to be estimated for both positive and negative images. Hence, it is beneficial to integrate  $S_{pix}$ with a  confidence score that indicates the presence of object $O$ in at least one pixel in the image. For this, we use the same  cSPP  image classifiers learnt earlier for each category.  The SVM associated with the cSPP image classifier gives a confidence score $\Phi(O) $ for a particular object $O$ as $\Theta $ $(Y=1\,|\,Z)$. 
		These scores are scaled between $0$  and $1$ as  
		\begin{equation}
		\hat \Phi(O)=\frac{exp(\Phi(O))}{\max\limits_{1\leq j\leq n_c}\{exp(\Phi(j))\}} \;,
		\end{equation}
		where,  $n_c$ is  the total number of categories. 
		Unlike soft-max that sums to 1, we normalize the score with the maximum  because multiple categories can simultaneously  appear in an image  such as in PASCAL VOC-2012~\cite{voc2012dataset}. In such scenarios, softmax will end up assigning a lower value to all positive categories. However, our objective is to identify the relative confidence across categories, and  assign $1$  to the most probable category. 
		To reduce false detections from less probable categories, we assume  values of  $\hat \Phi(O)$ that are less than $0.5$ as less important, and replace it with $ 0$. This limits the number of probable object categories per image to less than 5 categories in most images, and hence the category-specific saliency map $S_{pix}$ needs to be computed only for these few probable object categories.  
		We compute the classifier-weighted, category-specific score for each object $O$,
		\begin{equation}
		S_{categ}(O)=S_{pix}(O)\cdot \hat \Phi(O).
		\label{eq:S_categ}
		\end{equation}
		
		%
			\vspace{-0.1cm}	
		\subsubsection{Category-independent salient object detection}
		\label{sec:Applic_categInd}
		
		The proposed  category-specific TD saliency map $S_{categ}$  in Eq.~(\ref{eq:S_categ}) can be used to  compute the category-independent saliency value  $S_{ind}$,  by computing the maximum saliency value at each pixel (${x}$,${y}$) as 
		\begin{equation*}
		S_{ind}({x},\,{y})=\max\limits_{1\leq j\leq n_c}\{ S_{categ}(j)({x},\, {y})\}.
		\end{equation*} 
		Since the bottom-up information  is integrated to 
		$S_{categ}$ through  the  combined saliency map $\mathcal H$,  the   $S_{ind}({x},{y})$ gives an  accurate estimate of saliency maps under free-viewing condition.
		\vspace{-0.41cm}	
		\section{Experimental evaluation}
		\label{sec:Experiments}
		  \vspace{-0.098cm}
		
		We evaluate our weakly supervised saliency model on    Graz-02~\cite{Graz_02_dataset}, PASCAL VOC-2012,  PASCAL VOC-2007~\cite{dataset_voc2007} and PASCAL-S~\cite{PascalSdataset} datasets. Additionally, we use PASCAL VOC-2012 segmentation test set and validation set  to  evaluate weakly supervised semantic segmentation, Object Discovery dataset~\cite{unsupervisedCosegCVPR2013} to compare with semantic object selection and  co-segmentation approaches,   and validation set of  PASCAL VOC-2012 detection challenge to evaluate object localization and object detection performance. 
		
		
		Graz-02 dataset contains 3 object categories and a background category with 300 images per category. We split the images into training and testing sets following \cite{topdownCVPR2012},  i.e., 150 odd numbered images from each category are used for  training and 150 even numbered images from each category for testing.   
		PASCAL VOC-2012 is another challenging dataset with category-specific  annotations for 20 object categories. 
		It has 5717 training images and 5823 validation images for image classification/object detection challenge. 
		There are 1464 training images, 1449  validation images  and 1456 test images for segmentation challenge.  Our PASCAL VOC-2012  saliency models are trained using  5717  training images  for  image classification task. There are 210 test images in the segmentation challenge of PASCAL VOC-2007.  
		
		PASCAL-S is a widely used  dataset to evaluate  category-independent saliency models. It has 850 images picked from the validation set of PASCAL VOC-2010~\cite{pascal-voc-2010} segmentation images. Given the segmented objects in an image, the ground truth  salient objects are marked by twelve subjects under free-viewing condition.    
		We use Object Discovery dataset~\cite{unsupervisedCosegCVPR2013} to evaluate object segmentation. The dataset has three object categories,  namely airplane, car and horse. Apart from 100 test images per category, there are 461, 1206 and  779 additional images for airplane, car and  horse, respectively.  

	 \vspace{-0.35cm}
			\subsection{Analysis of proposed framework}
			 \vspace{-0.1cm}
			\begin{figure}[t]
				
				\includegraphics[width=1\linewidth, clip=true, trim= 0.0cm 11.0cm 5.5cm 0.5cm]{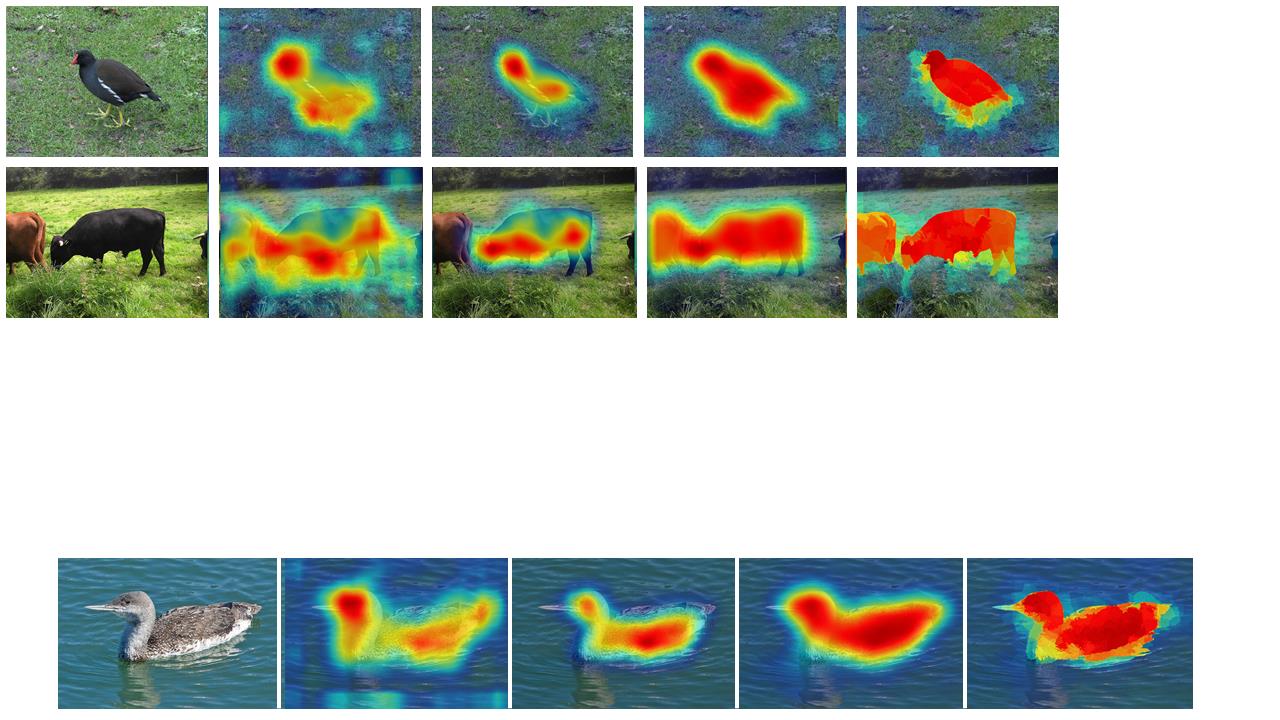}
				
				\hspace*{0.03\textwidth} (a) \hspace*{0.07\textwidth}(b)\hspace*{0.07\textwidth}(c)  \hspace*{0.07\textwidth} (d) \hspace*{0.07\textwidth} (e)	\vspace{-0.2cm} 
				\caption{Qualitative results at individual stages of the proposed method. (a) Input image, (b) B-cSPP saliency map, (c)  (b)~+~bottom-up saliency, (d) (c)\,+\,feature saliency, (e) (d)\,+\,superpixel averaging.}
				
				\label{fig:StepByStepEvaluationFig}
			\end{figure}
			
		\subsubsection{Contribution of individual modules}
		\label{sec:individualModuleCNN}
		Fig.~\ref{fig:StepByStepEvaluationFig} shows the visual comparison of the effect of each stage in the proposed method. For the input images in Fig.~\ref{fig:StepByStepEvaluationFig}(a), image regions  containing  bird's  head and cow's legs make positive contribution to their image classifiers and are, therefore,  assigned  high B-cSPP saliency in  Fig.~\ref{fig:StepByStepEvaluationFig}(b). Combining B-cSPP saliency with BU saliency  removed false detections in B-cSPP saliency as shown in Fig.~\ref{fig:StepByStepEvaluationFig}(c). Integration of feature saliency  assigns higher saliency value to the non-discriminative  object regions (Fig.~\ref{fig:StepByStepEvaluationFig}(d)). Finally, the addition of  the multi-scale superpixel-averaging improved the accuracy along object boundaries as shown in  Fig.~\ref{fig:StepByStepEvaluationFig}(e).

		\begin{figure}[t]
			\centering
			
			\fbox{\includegraphics[width=0.95\linewidth, clip=true, trim= 2.5cm 2.0cm 2.6cm 5.5cm]{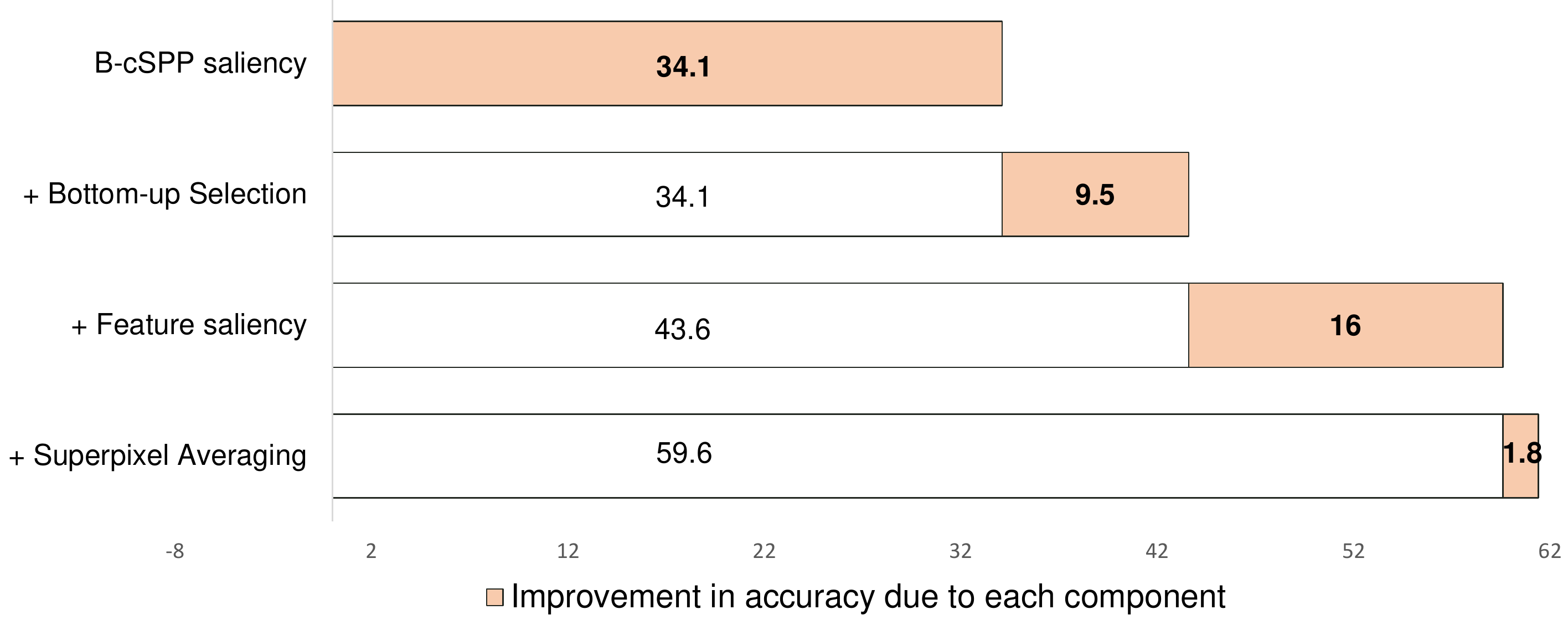}}\\ 
			\caption{Evaluation of individual stages of the proposed framework across 20 categories of PASCAL VOC-2012 using pixel-level precision rate at EER. The improvement in accuracy by the  addition of each module is shaded. }
			   \vspace{-0.3cm}
			\label{fig:EffectOfEachModulCNN}
		\end{figure}
		
			\begin{table*}[t]
				\centering
				\begin{bluetext} 
				\caption{Pixel-level precision rates at EER (\%) to analyze usefulness of BU components}
				\label{useofbottomUpExp}
				\begin{tabular}{|l|c|c|c|}
					\hline
					Evaluation of individual components & \begin{tabular}[c]{@{}c@{}}PASCAL VOC-2012\\  Segmentation  validation set\end{tabular} & \multicolumn{1}{l|}{Graz-02} & \multicolumn{1}{l|}{PASCAL-S} \\ \hline
					\begin{tabular}[c]{@{}l@{}}Without BU and without super pixel averaging.\\ Only top-down modules (B-cSPP+feature saliency)\end{tabular} & 50.2 & 69.1 & 42.8 \\ \hline
					\begin{tabular}[c]{@{}l@{}}With BU and without super pixel averaging.\\  (B-cSPP+BU+feature saliency)\end{tabular} & 59.6 & 78.1 & 55.0 \\ \hline
					\begin{tabular}[c]{@{}l@{}}Without BU and With super pixel averaging\\ (B-cSPP+feature saliency+superpixel averaging)\end{tabular} & 57.1 & 79.9 & 48.5 \\ \hline
					\begin{tabular}[c]{@{}l@{}}Full framework\\ (B-cSPP+BU+feature saliency+super pixel averaging)\end{tabular} & 61.4 & 82.5 & 67.5 \\ \hline
				\end{tabular}
					\end{bluetext}
			\end{table*}
		
		We evaluate the improvement in the mean precision rate (\%) at EER at each stage of our framework. The evaluation is done across 20 object categories  of  PASCAL VOC-2012 segmentation-validation set. 
		The contribution of each component in the proposed saliency model to the final accuracy  is shown  shaded  in  Fig.~\ref{fig:EffectOfEachModulCNN}. 
		The  accuracy of B-cSPP saliency is 34.1\%. On adding the BU map to yield combined saliency, the accuracy increased to 43.6\%, demonstrating the effectiveness of the proposed BU selection strategy. 
		
		Training the saliency model using negative patches from positive training images  improve the accuracy  by  5\%  in \cite{HishamCVPR2016}.  In the proposed framework, the accuracy of the combined saliency map {$\mathcal H$}  is  improved by weighting B-cSPP with the selected BU map, which enabled us to train  the feature saliency  using  negative patches from positive images.  This resulted in an additional improvement of 3\% in accuracy, totaling to  16\%  with the addition of feature saliency.  
		This demonstrates that (i) learning a dedicated feature classifier plays an important role for  TD saliency and (ii) combined saliency map $\mathcal H $ and feature  saliency map complement each other. A similar trend in improvement was observed in \cite{HishamCVPR2016}, where a feature classifier is learnt using contextual max-pooled  sparse codes. Since  relu5 features span larger spatial context compared to SIFT features computed on $64\times 64$ patches, contextual max-pooling on relu5 features is not required. 
		
		The feature saliency map $ \mathcal L$  and combined saliency map $\mathcal H$ are integrated  as $mean~(\mathcal H,~ \mathcal L )$. Other combinations such as  $max~(\mathcal H,~ \mathcal L )$ and  $\mathcal H \mathcal L+0.5(\mathcal H+\cal L)$ gave  similar results with less than  1\% variation in  accuracy.
		Taking the product of both saliency maps reduced the accuracy by 6\% as the combined saliency  $\mathcal H$  is often  0 in non-discriminative object regions and multiplication causes 0 values in such locations of the integrated map, disregarding  feature saliency.
		%
		
		Finally,   superpixel-averaging is applied at  6 different scales, by extracting   8, 16, 32, 64, 128 and 256 superpixels from an image. The saliency values at each pixel are further averaged across these 6 scales  
		to get the saliency map $S_{pix}$.    
		The quality of the saliency map at object boundaries is improved leading to  1.8\%  improvement in the accuracy,  to obtain  an  accuracy of 61.4\%.  Since superpixel computation at multiple scales is  time consuming relative to other modules, inference speed can be largely improved in applications such as object localization that do not require exact object boundaries  by  removing this step. 
		%
			\begin{bluetext} 		
		
		 \subsubsection{Evaluation of category-independent components}
		 \label{evalBUcomp}
		 There are two category-independent components in our framework: bottom-up selection and superpixel averaging.  Effectiveness of these components are evaluated in this section. We  evaluated the performance of top-down modules (B-cSPP, feature saliency)  by removing both BU selection and superpixel averaging modules.  From row 1 and row 2 of Table~\ref{useofbottomUpExp}, it can be observed that addition of BU component  improves the accuracy by 9\% in category-specific datasets Graz-02 and PASCAL VOC 2012, while  it has larger impact on category-independent dataset PASCAL-S (12.2\%).

		 Addition of multi-scale superpixel averaging to the framework with BU component helped to improve the accuracy by 1.8\%, 4.4\% and 12.5\% on PASCAL VOC 2012, Graz-02 and PASCAL-S datasets (row 2 vs row 4), respectively. These results shows that having category-independent components in the framework helps in improving both category-specific and category-independent saliency detection tasks and it has higher impact on category-independent saliency detection.

		 We further analyzed the performance of super-pixel averaging on a framework without BU, and the results indicate that   superpixel averaging has larger role in removing false positives in the top-down saliency detection through multi-scale averaging of saliency maps  
		 as seen by increased accuracy of 6.9\% and 10.8\% on PASCAL VOC 2012 and Graz-02 datasets respectively (row 1 vs row 3).   It is to be noted that the proposed bottom-up selection strategy  uses   fast and  training-free bottom-up saliency approaches and hence it is    faster compared to  super-pixel averaging (the other category-independent component in our framework) as explained in Section II  of the supplementary material. 

			   	
			   	
			  \end{bluetext}
			 
		\subsection{Comparison with other approaches}

			\begin{table}[t]
				\centering
				\caption{Pixel-level precision rates at EER (\%) on Graz-02.} 
				\vspace{-0.1cm} 
				\label{table:GrazWeakly150_600PixelCNN}
				\resizebox{0.98\linewidth}{!}{%
					\begin{tabular}{lccllll}
						\hline
						\textbf{Method}                   &   \multicolumn{1}{l}{SV}    & \multicolumn{1}{l}{\textbf{Test set}}                                                                                         & \textbf{Bike}                 & \textbf{Car}                  & \textbf{Person}               & \textbf{Mean}                 \\ \hline
						\multicolumn{1}{l|}{1 - Yang and Yang \cite{topdownCVPR2012}}                       & \multicolumn{1}{c|}{FS} & \multicolumn{1}{c|}{}                                                                                                         & 59.4                          & 47.4                          & 49.8                          & 52.2                          \\
						\multicolumn{1}{l|}{2 - Kocak \textit{et al.} \cite{topdownBMVC2014}}                        & \multicolumn{1}{c|}{FS} & \multicolumn{1}{c|}{}                                                                                                         & 59.9                        & 45.2                         & 51.5                         & 52.2                         \\
						
						\multicolumn{1}{l|}{3 - LCCSC \cite{HishamTopdownBMVC2015}}                               & \multicolumn{1}{c|}{FS} & \multicolumn{1}{c|}{}                                                                                                         &   69.1                            &     58.4                           &    58.2                           &          61.9                     \\
						\multicolumn{1}{l|}{4 - CG-TD \cite{HishamImageComm2016}}           & \multicolumn{1}{c|}{FS} & \multicolumn{1}{c|}{}                                                                                                         &  64.4                            &     50.9                          &    56.4                          &         57.2                    \\                   
						
						\multicolumn{1}{l|}{\cellcolor[HTML]{BBDAFF}5 - WS-SC\cite{HishamCVPR2016} } & \multicolumn{1}{c|}{WS} & \multicolumn{1}{c|}{}                                                                                                         & \cellcolor[HTML]{BBDAFF}64.0 & \cellcolor[HTML]{BBDAFF}45.1 & \cellcolor[HTML]{BBDAFF}55.2 & \cellcolor[HTML]{BBDAFF}54.8 \\
						\multicolumn{1}{l|}{\cellcolor[HTML]{BBDAFF}6 - Proposed}  & \multicolumn{1}{c|}{WS} & \multicolumn{1}{c|}{\multirow{-8}{*}{\begin{tabular}[c]{@{}c@{}}All\\ test\\ images\end{tabular}}}                            & \cellcolor[HTML]{BBDAFF}80.5  & \cellcolor[HTML]{BBDAFF}61.4  & \cellcolor[HTML]{BBDAFF}75	.0  & \cellcolor[HTML]{BBDAFF}72.3 \\ \hline
						
						\multicolumn{1}{l|}{7 - MB~\cite{BottomUpICCV2015MDB}}                           & \multicolumn{1}{c|}{TF} & \multicolumn{1}{c|}{}                                                                                                         & 54.67                         & 39.03                         & 52.04                         & 48.58                         \\
						\multicolumn{1}{l|}{8 - Aldavert \textit{et al.}~\cite{aldavert2010fast}}                     & \multicolumn{1}{c|}{FS} & \multicolumn{1}{c|}{}                                                                                                         & 71.9                          & 64.9                          & 58.6                          & 65.13                         \\
						\multicolumn{1}{l|}{9 - Fulkerson  \textit{et al.}~\cite{fulkerson2009class}}                    & \multicolumn{1}{c|}{FS} & \multicolumn{1}{c|}{}                                                                                                         & 72.2                          & 72.2                          & 66.1                          & 70.16                         \\
						\multicolumn{1}{l|}{10 - Shape mask~\cite{shapeMaskIJCV2012}}                          & \multicolumn{1}{c|}{FS} & \multicolumn{1}{c|}{}                                                                                                         & 61.8                          & 53.8                          & 44.1                          & 53.23                         \\
						\multicolumn{1}{l|}{11 - Yang and Yang \cite{topdownCVPR2012}}                       & \multicolumn{1}{c|}{FS} & \multicolumn{1}{c|}{}                                                                                                         & 62.4                          & 60                            & 62                            & 61.33                         \\
						\multicolumn{1}{l|}{12 - Khan and Tappen~\cite{icip2013discriminative}}                     & \multicolumn{1}{c|}{FS} & \multicolumn{1}{c|}{}                                                                                                         & 72.1                          & -                             & -                             & -                             \\
						\multicolumn{1}{l|}{13 - CG-TD \cite{HishamImageComm2016}}                               & \multicolumn{1}{c|}{FS} & \multicolumn{1}{c|}{}                                                                                   & 67.3                         & 59.8                          & 57.1        & 61.4                  \\
						\multicolumn{1}{l|}{\cellcolor[HTML]{BBDAFF}14 - WS-SC\cite{HishamCVPR2016} } & \multicolumn{1}{c|}{WS} & \multicolumn{1}{c|}{}                                                                                                         & \cellcolor[HTML]{BBDAFF}67.5  & \cellcolor[HTML]{BBDAFF}56.5 & \cellcolor[HTML]{BBDAFF}57.56 & \cellcolor[HTML]{BBDAFF}60.5 \\
						\multicolumn{1}{l|}{\cellcolor[HTML]{BBDAFF}15 - Proposed}  & \multicolumn{1}{c|}{WS} & \multicolumn{1}{c|}{\multirow{-10}{*}{\begin{tabular}[c]{@{}c@{}}Test \\ images\\ from\\ respective\\ category\\ \end{tabular}}} & 
						\cellcolor[HTML]{BBDAFF}84.1 & \cellcolor[HTML]{BBDAFF}81.5 & \cellcolor[HTML]{BBDAFF}81.8 & \cellcolor[HTML]{BBDAFF}82.5\\ \hline
					\end{tabular}
				}
				\vspace{-0.5cm}
			\end{table}
			
		\subsubsection{Graz-02 dataset}
		\label{sec:Graz-02 dataset}
		
		\begin{table*}[t]
			\centering
			\caption{Pixel-level precision rates at EER on validation set of PASCAL VOC-2012 segmentation dataset.The proposed weakly supervised approach outperforms all  fully supervised approaches including  \cite{ExmSal_CVPR2016}, which is based on CNN, in 14 out of 20 classes and in mean accuracy.}
			\label{table:VOC12Saliency}
			\resizebox{1\linewidth}{!}{%
				\begin{tabular}{l|l|llllllllllllllllllll|l}
					\hline
					Method & SV & plane & bike & bird & boat & botl & bus & car & cat & chair & cow & table & dog & horse & moto & pers & plant & sheep & sofa & train & tv & Mean \\ \hline \hline
					Yang \cite{topdownCVPR2012}  & FS &  14.7  & 28.1  &  9.8 &  6.1 &  2.2 &  24.1 &  30.2 &  17.3 &  6.2 &  7.6 &  10.3 & 11.5  & 12.5  & 24.1  &  36.7 &  2.2 &  20.4 & 12.3  &  26.1 & 10.2  &  15.6 \\
					Kocak \cite{topdownBMVC2014} & FS & 46.5 & \textbf{45.0} & 33.1 & \textbf{60.2} & 25.8 & 48.4 & 31.4 & 64.4 & 19.8 & 32.2 & 44.7 & 30.1 & 41.8 & 72.1 & 33.0 & 40.5 & 38.6 & 12.2 & 64.6 & 23.6 &  40.4   \\ 
					
					Exemplar \cite{ExmSal_CVPR2016} & FS &  55.9 & 37.9  & 45.6  &  43.8 &  47.3 &  \textbf{83.6 }&  57.8 &  69.4 &   22.7 &  68.5 &  37.1 &  72.8  &  63.7  &  69.0 & 57.5  &  \textbf{43.9} &  66.6 &  38.3 &  \textbf{75.1} & \textbf{56.7}  &  56.2\\ 
					
					Oquab \cite{LocOquab}& WS & 48.9 & 42.9 & 37.9  &47.1  & 31.4  & 68.4 & 39.9  & 66.2  & \textbf{27.2} & 54.0 & 38.3 &  48.5 & 56.5  & 70.1 & 43.2 & 42.6 & 52.2 & 34.8 & 68.1 & 43.4 &  48.1 \\  
					
					
					Proposed & WS & \textbf{71.2}   &  22.3  & \textbf{74.9}  &  39.9 &   \textbf{52.5}  &  82.7 &    \textbf{58.9 }&   \textbf{83.4} &    27.1  &  \textbf{81.1} &   \textbf{49.3} &  \textbf{82.4} &  \textbf{77.9} & \textbf{74.2}  & \textbf{69.8} &   31.9 &   \textbf{81.4}  &    \textbf{49.8} &    63.2 &    53.3 & \textbf{61.4}
					\\\hline   
				\end{tabular}%
			}
			\vspace{-0.4cm}
		\end{table*}

		We report our pixel-level results on different test set configurations of Graz-02.
		First, the proposed saliency model is compared with  other  TD saliency algorithms~\cite{topdownCVPR2012,topdownBMVC2014,HishamCVPR2016,HishamImageComm2016,HishamTopdownBMVC2015}   on all  600 test images. 
		Second, for comparison with related  approaches~\cite{shapeMaskIJCV2012,fulkerson2009class}, each object category is evaluated on  test images from its respective category. 
		Finally, to compare with \cite{topdownDSD,topdownSUN},  results on 300 test images are evaluated, where 150 test images are from a single category and the remaining 150 are from the background. 

		 The pixel-level comparisons  in the first two test set configurations are shown in Table~\ref{table:GrazWeakly150_600PixelCNN}, where  SV indicates supervision level with WS, FS and TF  referring to  weakly supervised, fully supervised and training-free approaches respectively. 
		\cite{topdownCVPR2012,topdownBMVC2014} and \cite{HishamImageComm2016}   are fully supervised (FS), needing multiple iterations of CRF learning with sparse codes relearned at each iteration. Separate dictionaries are used for each object category. On the contrary, the proposed weakly supervised method does not require any iterative learning and the relu5 features are extracted  with a single forward pass on the CNN. 
		\cite{HishamCVPR2016}  does not require any iterative learning and  uses a smaller dictionary of 1536 atoms,  compared to 2048 atoms used in  \cite{HishamTopdownBMVC2015}. Despite incorporating objectness~\cite{objectness} and superpixel features to \cite{topdownCVPR2012}, 
		the discriminative capability of \cite{topdownBMVC2014} did  not improve (row 2 vs row 1). 
		The  proposed weakly supervised  method (row 6 and row 15)  outperforms all other  fully supervised TD saliency approaches   \cite{topdownCVPR2012,topdownBMVC2014,HishamImageComm2016} .

		With respect to the second test configuration, \cite{shapeMaskIJCV2012}  requires images to be marked as \emph{difficult} or \emph{truncated} in addition to the object annotation for training of shape mask. 
		\cite{aldavert2010fast} uses 500,000 dictionary atoms in their fully supervised framework  to obtain 65.13\% (row 8), whereas the  dimension of our relu5 feature is  only 512. In this test setting, the proposed method achieves a mean accuracy of 82.5\%  outperforming the previous weakly supervised approach  \cite{HishamCVPR2016} by 22\%.


	    DSD~\cite{topdownDSD} has limited capability to remove background clutter,
	    resulting in poor performance of their model. DSD~\cite{topdownDSD} and SUN~\cite{topdownSUN} did not evaluate their model on Graz-02 dataset, but Yang and Yang \cite{topdownCVPR2012} reported their patch-level precision rates at EER on 300 test images as  49.4\% and 53.3\%,  respectively. 
		Feature learning using independent component analysis helped SUN to perform better than DSD, but substantially poorer than \cite{HishamCVPR2016} (65.4\%)  and the proposed method.  It is to be noted that the performance  of \cite{topdownDSD,topdownSUN,topdownCVPR2012,HishamCVPR2016}  deteriorates while converting  their  patch-level results to pixel-level.  
		The proposed weakly supervised method gives a mean pixel-level precision rate at EER of 73.1\%  which is better than the  70.16\% and 70.49\%  reported by \cite{topdownBMVC2014} and \cite{HishamTopdownBMVC2015} respectively in this test setting.  
		In all the three test settings, the proposed modifications enabled our current model to outperform \cite{HishamCVPR2016} by more than 18\% in accuracy, achieving state-of-the art performance.  The use of CNN features contributes mainly to this performance boost. We could not compare with~\cite{VisualizeCNN_ICLR2014} and  \cite{excitationBackPropECCV2016}, since they did not report their salient object detection results.  Qualitative comparisons with fully supervised and weakly supervised TD saliency approaches are shown in Fig.~\ref{fig:ComparisonImgFS} and in the supplementary material, respectively.  

		\subsubsection{PASCAL VOC-2012 segmentation dataset}

		In Table~\ref{table:VOC12Saliency}, we compare  a  CNN-based fully supervised TD saliency \cite{ExmSal_CVPR2016} with our method by evaluating on PASCAL VOC-2012 segmentation-validation set consisting of 1449 images. Similar to \cite{ExmSal_CVPR2016}, each object category is evaluated only on positive images of that category. We did  not fine-tune the convolution layers for this dataset, which took nearly 8 days on a GPU in \cite{ExmSal_CVPR2016}. 
		The presence of multiple, visually similar object classes in a single image is challenging for a weakly supervised approach. 
		Inspite of  this, we outperform the  state-of-the art fully supervised approach \cite{ExmSal_CVPR2016} and the CNN-based weakly supervised object localization approach \cite{LocOquab} in mean accuracy by 5\%  and 13\%, respectively. 
		We outperform \cite{LocOquab} in 15 out of the 20 categories.
		The top-down selection of BU approach along with feature saliency plays an important role in this improved performance, especially in classes like aeroplane and sheep. 
		
		\subsubsection{PASCAL VOC-2007 segmentation dataset}

		Following \cite{HishamTopdownBMVC2015,topdownBMVC2014}  and  \cite{topdownCVPR2012}, the saliency models are evaluated on 210 segmentation test images.  
		We used  the  models trained on PASCAL VOC-2012 training set in this experiment. Separate sparse codes of size 512 are computed for each category in ~\cite{topdownCVPR2012,topdownBMVC2014} and \cite{HishamImageComm2016}. \cite{HishamCVPR2016} uses sparse coding on a common dictionary of 1536 atoms for all object classes.
		Similarly, a common feature code of $20\times 512 = 10240 $ elements is used in \cite{HishamTopdownBMVC2015}. In our method, we compute 512 dimensional relu5 features which are common for all object categories.
		
		
		
		Table~\ref{PASCALAvgPrec} compares the pixel-level performance of the
		proposed WS method and patch-level results of FS top-down saliency approaches~\cite{topdownCVPR2012, HishamTopdownBMVC2015}  (these approaches did not report their  pixel-level results on this dataset). 
		We outperform \cite{topdownCVPR2012}, \cite{HishamImageComm2016} and \cite{HishamTopdownBMVC2015} in almost all categories and 
		in mean  precision rate at EER across 20 classes.
		A performance drop of 5 to 10\%  is reported by \cite{icip2013discriminative} while converting patch-level results of ~\cite{topdownCVPR2012} to pixel-level, which further increases the performance gain of the proposed approach. 
			\begin{table}[t]
				\centering
				\caption{Precision rates at EER(\%) on PASCAL VOC-2007.}
				\label{PASCALAvgPrec}
				\resizebox{0.85\linewidth}{!}{%
					\begin{tabular}{cccccc}
						\hline
						Method                                                        & \multicolumn{1}{c}{\begin{tabular}[c]{@{}c@{}}Yang and \\ Yang \cite{topdownCVPR2012}\end{tabular}} & \multicolumn{1}{c}{\begin{tabular}[c]{@{}c@{}}LCCSC~\cite{HishamTopdownBMVC2015}\\\end{tabular}} & \multicolumn{1}{c}{\begin{tabular}[c]{@{}c@{}}CG-TD \cite{HishamImageComm2016}\\\end{tabular}} & \begin{tabular}[c]{@{}c@{}}WS-SC\cite{HishamCVPR2016}\end{tabular} & Proposed \\ \hline
						\multicolumn{1}{l}{Supervision}      & FS          &   FS                                                                            & FS                                                                            & WS                                                                   & WS       \\ \hline
						\begin{tabular}[c]{@{}c@{}}Mean of\\  20 classes\end{tabular} & 16.7                                                                         & 23.4                                                                          & 23.81                                                                         & 18.6                                                                 & 42.1    \\ \hline
					\end{tabular}%
				}
					\vspace{-0.4cm}
			\end{table}
		Khan and Tappen~\cite{icip2013discriminative} report  pixel-level precision rates at EER  only for cow category (8.5\%) which is  much lower than the proposed weakly supervised approach~(52.3\%).

		\subsubsection{Category-independent salient object detection}
		\begin{figure}[t]
			\centering
			
			\fbox{\includegraphics[width=0.94\linewidth, clip=true, trim= 0.0cm 5.2cm 1.0cm 2.1cm]{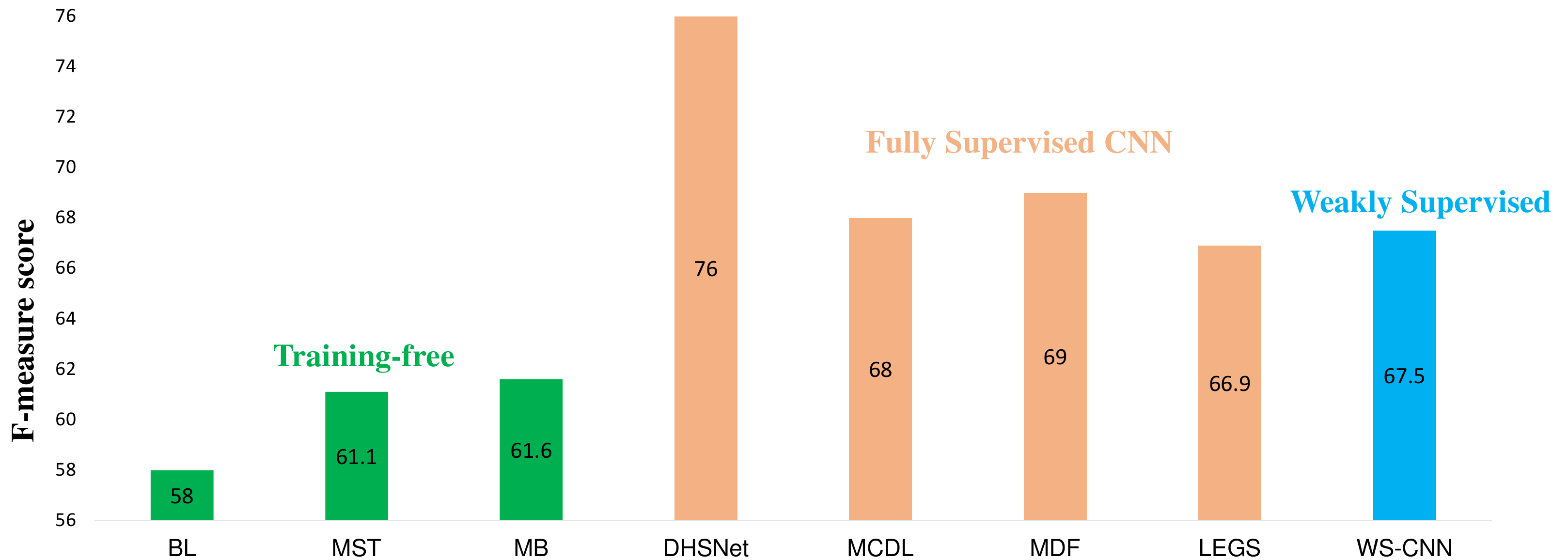}}\\ 
			\caption{Comparison of the proposed weakly supervised approach with state-of-the-art category-independent  saliency approaches on PASCAL-S dataset. We achieve a performance comparable with deep learning-based fully supervised approaches. }
			   \vspace{-0.5cm}
			\label{fig:PascalSImg}
		\end{figure}
		Category-independent saliency maps are obtained using the top-down models trained on PASCAL VOC-2012 training set through simple pixel-level maximum operation as explained in Section~\ref{sec:Applic_categInd}. Saliency values less  than 0.5 are considered as background, and those between 0.5 and 1 are normalized to $[0,\,1]$. The performance is evaluated on PASCAL-S dataset.   
		Fig.~\ref{fig:PascalSImg} compares the proposed method against state-of-the-art category-independent approaches that include deep learning based fully supervised approaches such as MCDL~\cite{MCDL}, LEGS~\cite{LEGS}, MDF~\cite{MDF} and DHSNet~\cite{DHSNet}. 
		The performance metric, F-measure is 
		\begin{equation*}
			\vspace{-0.02cm}
		f_{\eta}=\dfrac{(1+{\eta}^2)\cdot Precision\cdot Recall}{{\eta}^2 \cdot Precision+ Recall},
		\vspace{-0.02cm}
		\end{equation*}                         
		where ${\eta}^2=0.3$~\cite{PascalSdataset}. Following \cite{DHSNet}, precision and recall are computed by  binarizing  each saliency map at an image adaptive threshold, which is twice the average value of the saliency map.
		
		The proposed  weakly supervised method   achieves an f-measure of   67.5, which is comparable with fully supervised LEGS, MCDL and MDF.        
		We use only 5717 images from PASCAL VOC-2012 training set, which is much smaller  compared to the training data used by fully supervised approaches shown in Fig.~\ref{fig:PascalSImg}. 
		For example, DHSNet uses nearly 10,000 fully annotated images from multiple datasets such as MSRA 10K~\cite{SalObjBenchmark} and DUT-OMRON~\cite{bottomUp_manifold}.  Data augmentation is used  to further increase the number of training images. With less supervision and lesser training data, we  achieve a performance comparable with these fully supervised approaches. Qualitative results are present in the supplementary material.

			 \vspace{-0.35cm}
		\subsection{Applications}
			 \vspace{-0.05cm}
			\begin{figure}[t]
				\centering
				\includegraphics[width=0.95\linewidth, clip=true, trim=0.5cm 14.2cm 10.5cm 0.2cm]{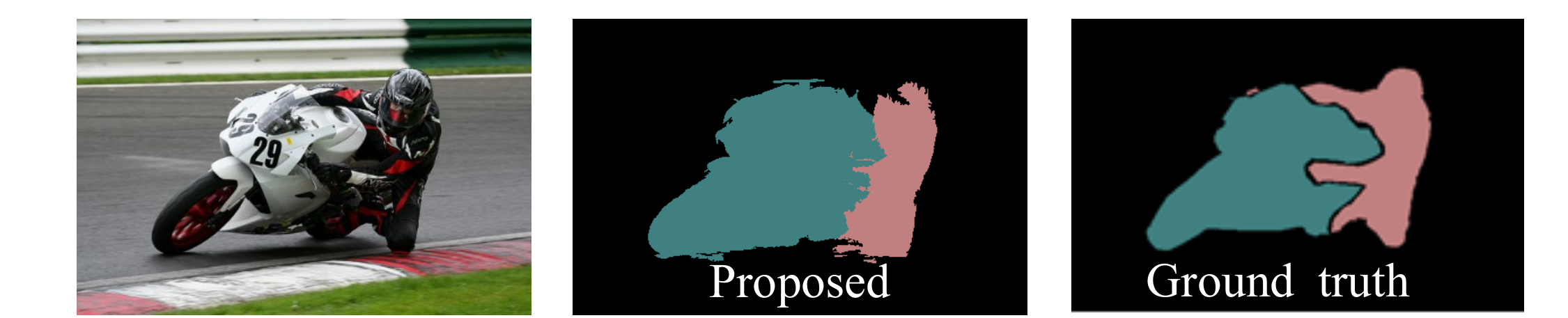} 
				\hspace*{0.0cm}(a) Semantic segmentation\\
				\vspace{0.2cm}
				\includegraphics[width=1\linewidth, clip=true, trim=0.5cm 9.5cm 10.5cm 4.9cm]{images_tip/TIP_applications1.pdf}
				\hspace*{0.0cm}(b) Object segmentation\\
				\vspace{0.2cm}
				\includegraphics[width=1\linewidth, clip=true, trim=0.5cm 4.8cm 10.5cm 9.6cm]{images_tip/TIP_applications1.pdf}
				\hspace*{0.0cm}(c) Object localization\\
				\vspace{0.2cm}
				\includegraphics[width=1\linewidth, clip=true, trim=0.5cm 0.1cm 10.5cm 14.3cm]{images_tip/TIP_applications1.pdf}
				\hspace*{0.0cm}(d) Object detection\\
				\vspace{0.2cm}
				
				\vspace*{-0.1cm}
				\caption{Applications of proposed saliency map for (a) Semantic segmentation, (b) object segmentation, (c) Object localization and (d) object detection.}
				\label{fig:ApplicationsTIP}
				 \vspace*{-0.3cm}
			\end{figure}
			
		TD saliency \cite{JimeiPAMI2016,topdownBMVC2014,HishamCVPR2016,ExmSal_CVPR2016} and   tasks like object detection, localization and segmentation mainly differ in their granularity of representation. 	
		Object detection  produces  a tight rectangular bounding box around all instances of objects belonging to   user-defined categories. It is necessary to identify both the location as well as the extent of each object. 
		The process of identifying the location of a particular object in an image, without  marking the  extent  of the object, is referred  to  as object localization~\cite{LocOquab}.  
		Object segmentation, also referred to as semantic object selection   produces a binary mask with `1' indicating all pixels that belong to a user-defined object category. It differs from the task of semantic segmentation, where the objective is to classify each pixel in the image to one of  predefined classes.			
		 In this section, we evaluate the use of our TD saliency framework for  various applications in a weakly supervised setting. 
		\subsubsection{Weakly supervised semantic segmentation}

%
			\label{sec:semantiSeg}	  
			The category-specific saliency maps in the proposed framework can be easily adapted for semantic segmentation. 
			In the saliency map, a pixel with $S_{categ}(O)<0.5$ is less likely to belong to an object $O$. The pixels at which the maximum saliency  across all categories is less than 0.5 is more likely to be background. Hence, an additional map corresponding to the background  category is generated as a uniform map with $S_{categ}=0.5$. We assign to each pixel the category for which its saliency is the maximum. 		
		
		In PASCAL VOC-2012 semantic segmentation task, each pixel in the image needs to be classified  to one of  21 categories comprising background  and 20 object categories. Our results are reported from PASCAL VOC-2012 evaluation server which uses intersection-over-union (IoU) ) as the evaluation metric.  The mean intersection-over-union (mIoU) on the validation set and test set are reported in  Table~\ref{Table:SemanticSeg_Valid}.  \textcolor{black}{The proposed approach requires the least number of training images (5k), yet it achieves a performance comparable with the state-of-the-art approaches~\cite{qi2016augmented,kolesnikov2016seed,saleh2016built,shimodaECCV16,wei2016stc} that use 10k to 50k training images.  	
		Unlike ~\cite{qi2016augmented,kolesnikov2016seed,saleh2016built,shimodaECCV16}, we do not use CRF or Grab-cut~\cite{rother2004grabcut}  or similar energy minimization techniques	to post-process the segmentation result}.

		Fig.~\ref{fig:ApplicationsTIP} compares the qualitative results obtained by the proposed method against the ground truth.  
		Majority of the person  pixels  are  classified correctly, despite the size of person being  small compared to motorbike in the top row. 
		\begin{table}[t]
			\caption{mean Intersection over union (mIOU) for semantic segmentation on  validation  and test sets of PASCAL VOC-2012.}
			\centering
			\label{Table:SemanticSeg_Valid}
			\resizebox{0.82\linewidth}{!}{%
				\begin{tabular}{@{}lccc@{}}
					\toprule
					\multicolumn{1}{c}{Method} & \begin{tabular}[c]{@{}c@{}}Training\\  set\end{tabular} & \begin{tabular}[c]{@{}c@{}}mIoU on\\  val set\end{tabular} & \begin{tabular}[c]{@{}c@{}}mIoU on\\  test set\end{tabular} \\ \midrule
					MIL-FCN~\cite{FullyconvICLR2015}       & 10k          & 25.7            & 24.9             \\
					CCNN~\cite{SemanticSeg_pathak}           & 10k          & 35.3            & 35.6             \\
					MIL-sppxl~\cite{semanticSegCVPR2015}    & 700k         & 36.6            & 35.8             \\
					MIL-bb~\cite{semanticSegCVPR2015}  & 700k         & 37.8            & 37.0             \\
					EM-Adapt~\cite{WeaklyPapandreou2015ICCV}       & 10k          & 38.2            & 39.6             \\
					STC \cite{wei2016stc}   & 50k          & 49.8            & 51.2             \\
					DCSM w/o CRF \cite{shimodaECCV16}  & 10k          & 40.5            & 41.0             \\
					DCSM w/ CRF  \cite{shimodaECCV16}  & 10k          & 44.1            & 45.1             \\
					BFBP w/ CRF \cite{saleh2016built}  & 10k          & 46.6            & 48.0             \\		
					SEC  w/ CRF  \cite{kolesnikov2016seed}  & 10k          & 50.7            & 51.7             \\		
					AF-SS w/ CRF~\cite{qi2016augmented} & 10k          & 52.6            & 52.7             \\\midrule
					Proposed                   & \textbf{5k}           & 43.5            & 44.1             \\ \bottomrule
				\end{tabular}
			}
				\vspace{-0.4cm}
		\end{table}

		\subsubsection{Weakly supervised object segmentation}
%
		
			\textcolor{black}{Conventional object segmentation approaches use scribbles or rectangular boxes to indicate the object of interest, while in our approach, only the semantic label of the object of interest is input to the system, similar to the semantic object selection~\cite{semanticObjectSeletion}. 
			We threshold our TD saliency map to  identify definite foreground and background regions in an image, followed by Grab-cut~\cite{rother2004grabcut}  to accurately segment out the object of interest. Being a weakly supervised approach, framework is comparable to   co-segmentation approaches that segment out a common object from a given set of images. We learn a model for the common object, which helps to achieve faster inference for a newly added test image, whereas  co-segmentation approaches need to re-segment every image in the  set upon encountering a new image}.
		
		Object segmentation accuracy is evaluated on 100 test images from each category of  Object Discovery  dataset~\cite{unsupervisedCosegCVPR2013}. 300 images from each category are used to train our saliency model,  along with 300 negative images from Graz-02 dataset.   
		Qualitative results are shown in Fig.~\ref{fig:ApplicationsTIP}.   Multiple instances of car  are accurately  segmented out and the proposed approach could accurately segment out the horse. 
		Quantitative comparisons with state-of-the-art co-segmentation approaches are shown in Table~\ref{table:coSegTableCNN}. The Jaccard similarity, i.e, intersection over union ($IOU$) with the ground-truth  is evaluated as in ~\cite{unsupervisedCosegCVPR2013}. 
		In all the three categories, we  achieve state-of-the-art performance compared to  related co-segmentation~\cite{unsupervisedCosegCVPR2013,Coseg_Quan_2016_CVPR} and co-saliency ~\cite{CoSalIJCV016} approaches. The  semantic object selection  \cite{semanticObjectSeletion} uses additional supervision  by
		collecting positive training images with white background using an internet search.  Inspite of this modification, they could only achieve an average accuracy of 63.73\%, which is lower than our mean accuracy of 68.0\% across 3 categories.
		%
			\begin{table}[t]
			\centering
			\caption{Comparison of proposed weakly supervised approach with object segmentation approaches on Object Discovery dataset, evaluated using Jaccard similarity.}
			\label{table:coSegTableCNN}
			\resizebox{0.82\linewidth}{!}{%
				\begin{tabular}{@{}lcccc@{}}
					\toprule
					Method          & Airplane & Car   & Horse   & Mean  \\ \midrule
					Joulin~\etal~\cite{joulin2010Cosegmentation}   & 15.4    & 37.2 & 30.2 & 27.6\\
					Joulin~\etal~\cite{multiclassCosegCVPR2012}   & 11.7    & 35.2 & 29.5 &25.5 \\
					Kim~\etal \cite{kim2011distributed}      & 7.9      & 0.04  & 6.43  & 4.79\\
					Object Discovey~\cite{unsupervisedCosegCVPR2013} & 55.8    & 64.4 & 51.6 & 57.3\\
					Koteshwar~\etal~\cite{Kot_TMM2016} & 56    & 69 & 55 & 60\\
					Zhang~\etal  \cite{CoSalIJCV016} & 53.5   & 58.8 & 52.2 & 54.8\\
					Quan~\etal \cite{Coseg_Quan_2016_CVPR} & 56.3    & 66.8 & 58.1& 60.4 \\
					WS-SC\cite{HishamCVPR2016}        & 57.3    & 67.4 & 50.51 & 58.4 \\ 
					Object selection \cite{semanticObjectSeletion} & 64.3 & 71.8 & 55.1& 63.7 \\ \midrule 
					Proposed        & \textbf{65.0}    & \textbf{77.3} & \textbf{61.6}& \textbf{68.0} \\ \bottomrule
				\end{tabular}
			}
				\vspace{-0.2cm}
		\end{table}
		
		\begin{table}[t]
			\centering
			\caption{Average precision of object localization on PASCAL VOC-2012 detection validation set. }			
			\label{table:LocalizationRes}
				\resizebox{0.82\linewidth}{!}{%
			\begin{tabular}{@{}lccc@{}}
				\toprule
				\multicolumn{1}{c}{Method}                & SV & Exact         & 18 Pixel      \\ \midrule
				RCNN~\cite{R-CNN}                         & FS & 67.7          & 74.8          \\
				Fast RCNN \cite{FastRCNN}                 & FS & -             & 81.9          \\
				Exemplar \cite{ExmSal_CVPR2016}           & FS & 73.4          & -             \\
				Oquab~\etal~\cite{LocOquab},              & WS &     -          & 74.5          \\
				ProNet~\cite{pronetCVPR2016}              & WS & 69.8          & 74.8          \\
				ProNet + classifier~\cite{pronetCVPR2016} & WS & 73.1          & 77.7          \\
				Bency \etal,\cite{Bency2016}              & WS & -             & 79.7          \\ \midrule 
				Proposed                                  & WS & \textbf{82.2} & \textbf{84.7} \\ \bottomrule
			\end{tabular}
		}
			\vspace{-0.4cm}
		\end{table}

		\subsubsection{Weakly supervised object localization}
		\textcolor{black}{Object localization deals with  locating  object $O$  within a positive image. Here, only the location  of the object needs to be identified, not its extent. The peaks of our saliency map, $S_{pix}$ indicates the location of object $O$,
			\begin{equation*}
			Loc~(O)=\underset{({x},\;{y})}{\operatorname{argmax}} ~\{ S_{Pix}(O)({x},\;{y})\}.
			\end{equation*}}
			
%
			
		Presence of multiple objects in an image makes object localization on PASCAL VOC-2012 detection set a challenging task, especially in a weakly supervised setting. Since an  accurate estimate of object boundaries are not required,  we replaced the multi-scale superpixel averaging with an averaging filter on a rectangular window of size  $64\times 64$ pixels for faster inference.  
		
		The location that  falls exactly within any ground truth bounding   box associated for a given category is assumed correct and the average precision is calculated as in  \cite{ExmSal_CVPR2016}. In \cite{LocOquab}, average precision is  evaluated by giving an error tolerance of 18 pixels  to the predicted location. We evaluated our model in both these settings denoted Exact and 18 Pix 
		and corresponding  results are compared with state-of-the art approaches  as shown  in Table~\ref{table:LocalizationRes}. 
		  In both the evaluation settings, we achieve 
		a performance which is  comparable to  fully supervised TD saliency approaches and dedicated object detectors such as fast RCNN~\cite{FastRCNN}.  It is to be noted that object localization is inherently a simpler task  compared to  object detection. In object detection  applications, fully supervised object detectors~\cite{fasterRCNN,YOLO,FastRCNN} outperform the proposed method, as expected. 
		Fig.~\ref{fig:ApplicationsTIP} shows some qualitative results obtained using the proposed method in localizing multiple objects. Partially occluded objects such as motorbike and car  are  localized accurately despite   the presence of other distracting objects. 
		
				\subsubsection{Weakly supervised object detection}
				
			\textcolor{black}{In object detection, multiple instances of the same object category need to be identified separately.  This is  more challenging than localization and especially so  in a weakly supervised setting. 
			Conventional object detectors such as R-CNN~\cite{R-CNN,ObjDetectCNNBilen} need to classify thousands of  category-independent object proposals generated using  selective search~\cite{SelectiveSearch}, \cite{EdgeProposal}. This  incurs a huge computational cost. 
			The proposed TD saliency  framework simplifies object detection by generating less than 5  proposals for an object category per image. 
			First, the category-specific saliency $S_{categ}(O)$ is binarized by applying a  threshold at $0.5$. The smallest rectangular box enclosing each disconnected region is the detection box for object $O$.  
			With this simple strategy, we achieve a performance which is  comparable to  dedicated weakly supervised object detectors~\cite{pronetCVPR2016}}.
		\begin{table}[t]
			\centering
			\caption{Comparison with weakly supervised object detection approaches on PASCAL VOC-2012 validation dataset, measured by average precision.}
			\label{table:detectionCNN}
			\resizebox{0.9\linewidth}{!}{%
				\begin{tabular}{@{}ccccc@{}}
					\toprule
					Method                                                                             & Oquab~\etal~\cite{LocOquab}  & ProNet~\cite{pronetCVPR2016}  & ProNet+Classifier~\cite{pronetCVPR2016} & Proposed \\ \midrule
					\begin{tabular}[c]{@{}c@{}}mAP\\ (Mean of 20 Classes)\end{tabular} & 11.74      & 13  & 15.5                & \textbf{20.4}         \\ \bottomrule
				\end{tabular}
			}
			   \vspace{-0.4cm}
		\end{table}
%

		The object detection boxes produced by  a simple binarization of our saliency maps is shown to be comparable with dedicated weakly supervised object detectors in  Table~\ref{table:detectionCNN}. 
		We outperform \cite{pronetCVPR2016} which uses  an additional box classifier to classify their object proposal boxes.        
		We consider all category-specific object  boxes as positive detections.  
		PASCAL VOC 2012 evaluation server is used to estimate object detection accuracy, where  
		a  detection  having an $IOU>0.5$ with the ground truth rectangular bounding box is  considered as true positive.    \cite{WSObjectDetection_DomainAdapt} requires thousands of forward passes  through the network to identify the class-specific object proposals, which is time consuming. Moreover, fine-tuning of  convolution layers is required for detection task, whereas we use the CNN trained for image classification, without fine-tuning. Inspite of all these computational requirements, their framework based on AlexNet  achieved a mAP of 22.4\% and their  framework based on \cite{VGGnet} achieved  a mAP of 29.1 on PASCAL VOC 2012 object detection dataset. 
		In Fig.~\ref{fig:ApplicationsTIP},  multiple overlapping objects are accurately detected by the proposed strategy. Multiple 
		instances of person, motorbike and  car  are also detected. 
		The person and bicycle are accurately detected despite the presence of other categories  in the image.  Similarly, an accurate bounding box around the cat is marked in an image that also contains a TV monitor. 
			
		
		
		\textbf{Limitations.}
		Similar to other weakly supervised approaches \cite{pronetCVPR2016,ObjDetectCNNBilen}, the proposed approach has   limited  ability to discriminate among multiple instances of an object which are spatially adjacent. This  causes  low performance for object detection, compared to state-of-the-art fully supervised object detectors~\cite{YOLO,FastRCNN,fasterRCNN}. Examples are provided in the supplementary material.
				
			\textbf{Supplementary material.}
		Additional experiments to evaluate (i) selection of bottom-up saliency map~\cite{bottomup_GP,extendedQCut}, (ii) performance comparison on different CNN architectures,  (iii) computation time, and (iv) additional qualitative  results and comparisons are included in the supplementary material. 
		\section{Conclusion}
		\label{sec:conclusion}  
		In this paper, a CNN feature-based weakly supervised salient object detection approach is proposed. 
		A novel strategy to select a BU saliency map that suits a top-down task is proposed. Contribution of relu5 features at different spatial locations are estimated to compute a novel B-cSPP saliency. The top-down B-cSPP saliency is integrated with the BU saliency map and produces a combined saliency which is further integrated  with feature saliency.
		 The proposed  weakly supervised top-down saliency  model  achieves state-of-the-art performance in top-down salient object detection across multiple datasets, by  outperforming  even fully supervised CNN-based approaches. Moreover, the top-down saliency maps of different object categories are combined to  produce  a category-independent saliency map  that can estimate salient objects under  free-viewing condition. Finally, through quantitative comparisons, we  demonstrated  the usefulness of proposed saliency map for four  different applications. We plan to extend our framework to videos for weakly supervised salient object detection. 
		{
			\bibliographystyle{IEEEtran}
			\bibliography{IEEEabrv_short,IEEEexample_new1}
		}

		\vfill
		

	\end{document}